\documentclass[11pt]{article}

\usepackage[final]{acl}

\usepackage{times}
\usepackage{latexsym}

\usepackage[T1]{fontenc}

\usepackage[utf8]{inputenc}

\usepackage{microtype}

\usepackage{inconsolata}

\usepackage{graphicx}

\usepackage[utf8]{inputenc}
\usepackage{amsmath, amssymb}
\usepackage{graphicx}
\usepackage{hyperref}
\usepackage{cite}

\usepackage{subcaption}  
\usepackage{caption}     

\usepackage{hyperref}

\usepackage[T1]{fontenc}
\usepackage{xcolor}

\usepackage{listings}
\usepackage{tcolorbox}
\tcbuselibrary{listings,skins,breakable}
\usepackage{tabularx}
\usepackage{booktabs}   
\usepackage{multirow} 
\usepackage{makecell}   
\usepackage{array}    
\usepackage{tabularx}
\usepackage{threeparttable}
\usepackage{siunitx}
\usepackage{fp}
\usepackage{caption}
\usepackage{listings}
\usepackage{xurl}  
\usepackage{booktabs}   
\usepackage{siunitx}    

\usepackage[symbol]{footmisc} 

\newcolumntype{Y}{>{\raggedright\arraybackslash}X}
\newcolumntype{C}[1]{>{\centering\arraybackslash}p{#1}}

\title{The Reasoning Lingua Franca: A Double-Edged Sword for Multilingual AI}

\author{
 \textbf{Alan Saji\textsuperscript{1\footnotemark[3]}},
 \textbf{Raj Dabre\textsuperscript{1,2,3}}
 \textbf{Anoop Kunchukuttan\textsuperscript{1,4}},
 \textbf{Ratish Puduppully\textsuperscript{5}\footnotemark[3] } \\
 \\
\textsuperscript{1}Nilekani Centre at AI4Bharat,
 \textsuperscript{2}Indian Institute of Technology Madras, India,
 \\
 \textsuperscript{3}Google,
\textsuperscript{4}Microsoft, India,  \textsuperscript{5}IT University of Copenhagen
}

\begin{document}
\maketitle
\begin{abstract}
Large Reasoning Models (LRMs) achieve strong performance on mathematical, scientific, and other question-answering tasks, but their multilingual reasoning abilities remain underexplored. When presented with non-English questions, LRMs often default to reasoning in English, raising concerns about interpretability and the handling of linguistic and cultural nuances. We systematically compare an LRM’s reasoning in English versus the language of the question. Our evaluation spans two tasks: MGSM and GPQA Diamond. Beyond measuring answer accuracy, we also analyze cognitive attributes in the reasoning traces. We find that English reasoning traces exhibit a substantially higher presence of these cognitive behaviors, and that reasoning in English generally yields higher final-answer accuracy, with the performance gap increasing as tasks become more complex. However, this English-centric strategy is susceptible to a key failure mode - getting \textbf{“Lost in Translation,"} where translation steps lead to errors that would have been avoided by question's language reasoning. Code is available at \url{https://github.com/AI4Bharat/multilingual_reasoning_analysis}.
\end{abstract}
\section{Introduction}



Large reasoning models (LRMs) \citep{guo2025deepseek,yang2025qwen3,jaech2024openai} have become popular over the past year thanks to their superior ability to analyze and answer questions across a range of tasks. Unlike traditional Large Language Models (LLMs) \citep{touvron2023llama,team2024gemma}, LRMs have a two-phase response generation. First, they generate a \textit{reasoning sequence}, which resembles a human's step-by-step reasoning process. This phase allows an LRM to do in-depth analysis, explore potential solutions and verify intermediate steps. After completing this reasoning stage, they generate an \textit{answering sequence} that succinctly presents the final answer derived from the reasoning sequence.

\footnotetext[3]{\small{
\textbf{Correspondence:} Alan Saji (\href{mailto:alansaji2001@gmail.com}{alansaji2001@gmail.com})\\,  Ratish Puduppully
(\href{mailto:rapu@itu.dk}{rapu@itu.dk})
 }}

LRMs tend to generate their reasoning sequences predominantly in English \citep{yong2025crosslingual}. For multilingual tasks, ideally the reasoning should be generated in the question’s language: this improves interpretability, lets users follow the model’s chain of thought directly, and preserves cultural and linguistic nuance \citep{aggarwal2025language}. Motivated by these considerations, we ask: \textit{how does reasoning in English compare to reasoning in the question’s language across multilingual tasks?}


In this study, we evaluate the 
above question on two benchmark datasets, MGSM \citep{shi2022language} and GPQA Diamond, which vary in difficulty \citep{rein2024gpqa} and together provide a comprehensive assessment of LRMs' reasoning capabilities. We analyze results from three perspectives: (1) which reasoning language yields correct final answers more frequently; (2) which reasoning sequences exhibit richer cognitive behaviors; and (3) in which conditions reasoning in the question’s language outperforms reasoning in English.

\begin{figure*}[t]
  \centering
  \includegraphics[width=\textwidth]{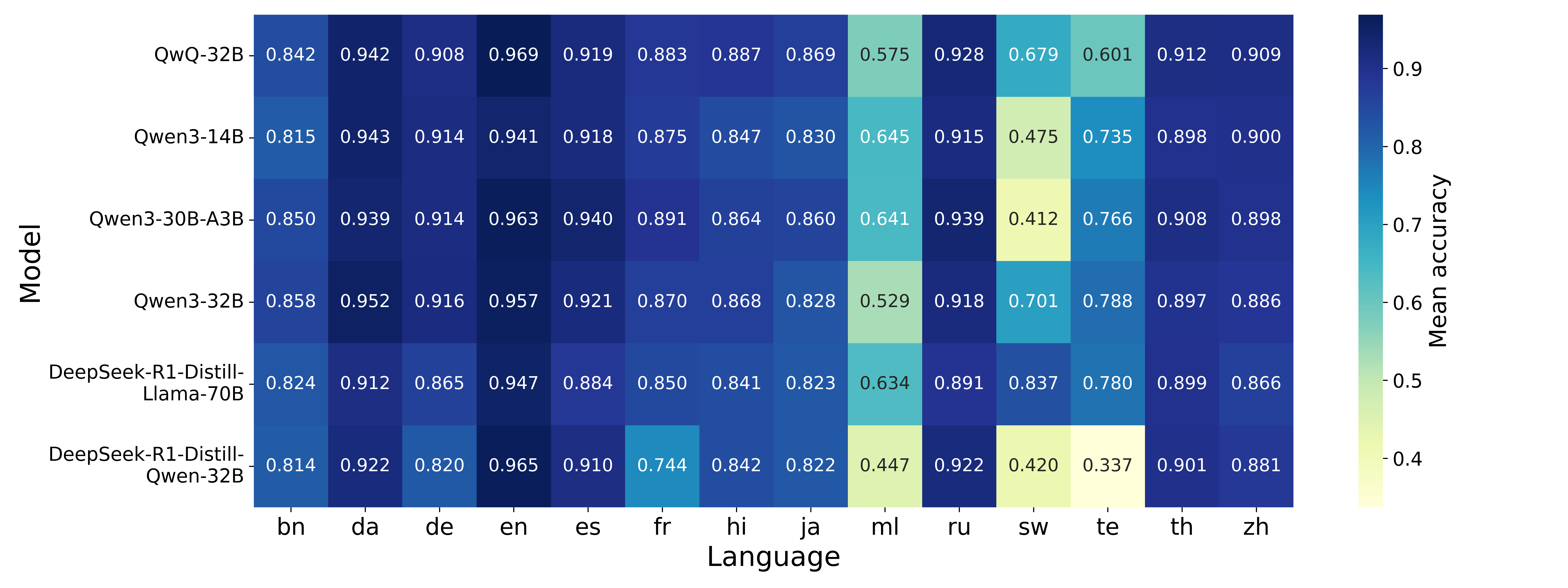}
    \caption{\textbf{MGSM - Reasoning in the Question’s Language.} Final answer accuracy of LRMs on MGSM when reasoning in question's language. Y-axis: models; X-axis: languages; each cell reports the avg@4 accuracy. Standard deviations are below 0.05 (c.f. Appendix \ref{sec: native_cot_reasoning_std}).}
    \label{fig:native_cot_reasoning}
\end{figure*}

\section{Related Work}
\label{related_work}

Recent studies have contrasted reasoning in English with reasoning in non-English languages \citep{wang2025polymath,yong2025crosslingual,languagemattersappier,qi2025models}. Our study extends this line of inquiry by quantifying how much reasoning in the question’s language diverges from reasoning in English across tasks of varying difficulty. In addition, we examine the distribution of cognitive behaviors within the reasoning traces for both settings. Finally, we highlight scenarios where reasoning in the question’s language can actually surpass English reasoning, providing insights that complement prior work.
See Appendix \ref{sec: Related Work - additional details} for more related work.
\section{Methodology}
\label{headings}

We compare how an LRM reasons in English versus reasoning in the language of the question for multilingual tasks, focusing on differences in final answer accuracy across language. Using a system prompt, the model is explicitly instructed to present its final answer within a \verb|\boxed{}| format, which is then used for answer extraction. To encourage models to reason in the question's language, we employ a system prompt (c.f. Appendix \ref{sec: sys prompt details}) and prepend prefix tokens to the reasoning trace (c.f. Appendix \ref{sec: Prepending Prefix tokens}), an approach shown to be effective in prior work \citep{yong2025crosslingual}.






\section{Experimental Settings}
The study employs MGSM \citep{shi2022language} and GPQA-Diamond \citep{rein2024gpqa} benchmark datasets, the latter translated into Danish and five Indic languages(c.f. Appendix \ref{sec:dataset}). We utilize open-weight reasoning models like Qwen3 32B \citep{yang2025qwen3} and DeepSeek-R1-Distill-Llama-70B \citep{guo2025deepseek} (c.f. Appendix \ref{sec: models}).

For LRMs, sampling-based decoding is generally favored over greedy decoding. We adopt the conventional hyperparameter setting of \(T=0.6\), \(p=0.95\), and \(k=20\) \citep{wang2025polymath} for LRMs. Each model is evaluated four times using these fixed hyperparameters, and we report both the mean and standard deviation across the four runs. We performed API-based inference using the Deepinfra API \citep{DeepInfraAPI2025}. Additional details about experimental settings, API cost and system prompts are present in Appendix~\ref{sec: Experimental setting additional details}.
Languages involved in this study are described in Appendix \ref{sec:Language Involved}.

\subsection{Grouping Languages by Resource Availability}
\label{sec: Grouping Languages by Resource Availability}
Among the languages analyzed in this study, English, Chinese, French, German, Japanese, Russian, Danish and Spanish are generally considered high-resource languages, as they have abundant training data, digital content, and linguistic resources available for large language models. In contrast, Hindi, Bengali, Malayalam, Gujarati, Telugu, Swahili, and Thai are typically regarded as low-resource languages, with comparatively limited high-quality text corpora and fewer NLP tools available. Within this group, languages such as Hindi and Bengali can be viewed as relatively better resourced (mid-resource) than others like Malayalam, Gujarati, Swahili, and Telugu, which remain low resource.

\begin{figure*}[htbp]
    \centering
        \hfill
    \begin{subfigure}{0.48\textwidth} 
        \centering
        \includegraphics[width= \textwidth]{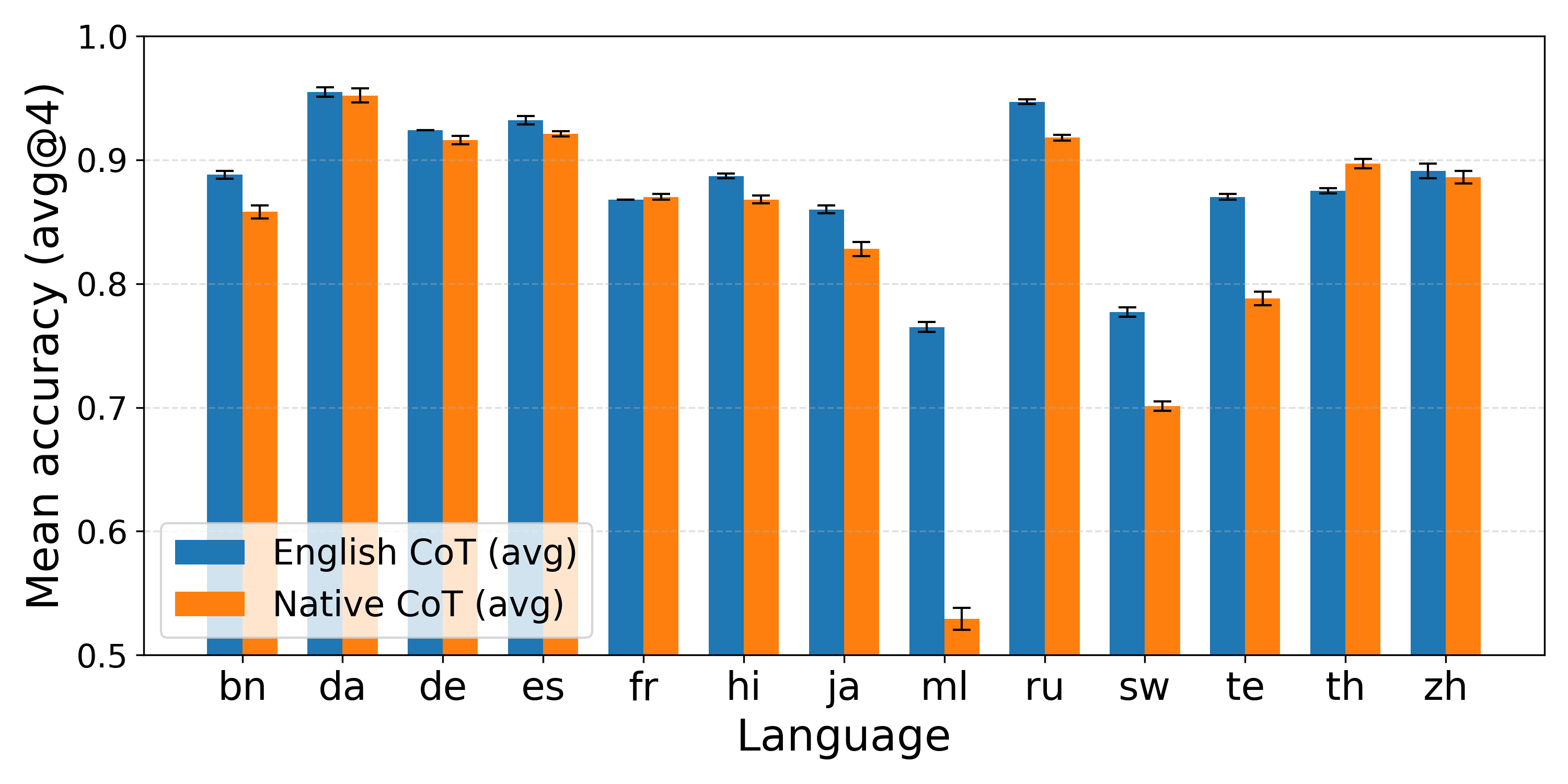} 
        \caption{MGSM}
        \label{fig:Qwen3-32B MGSM}
    
    \end{subfigure}
     \hfill
    \begin{subfigure}{0.48\textwidth} 
        \centering
        \includegraphics[width= \textwidth]{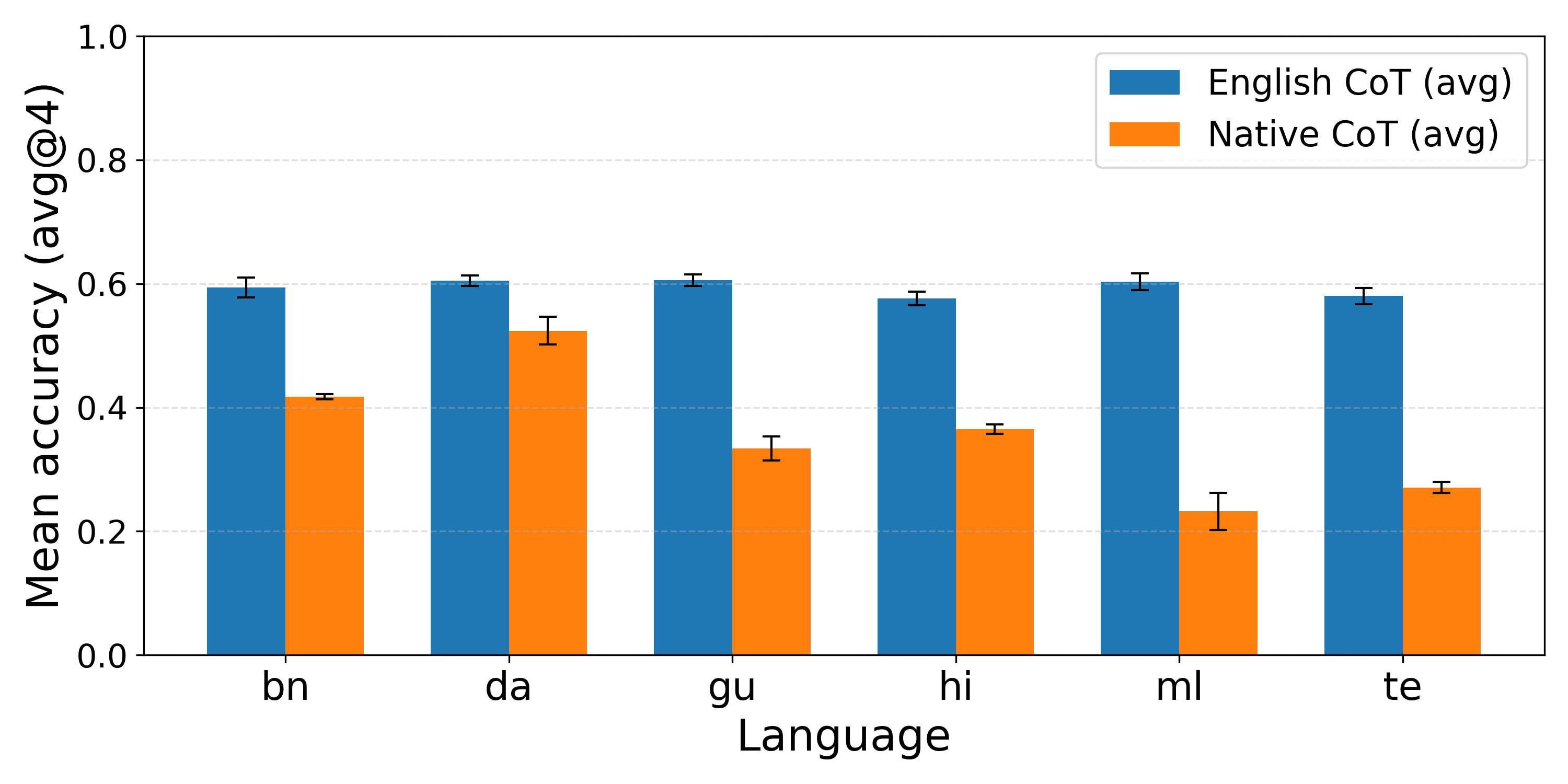} 
        \caption{GPQA Diamond}
        \label{fig:Qwen3-32B GPQA}
    \end{subfigure}
     \hfill
     \caption{\textbf{Reasoning in English vs Question's Language} is compared via final answer accuracy for \textbf{Qwen3 32B} for MGSM and GPQA diamond task. Error bars denote standard deviation.}
                
\label{fig:en_cot_vs_native_cot}
\end{figure*}

\begin{figure*}[t]
  \centering
  \includegraphics[width= \textwidth]{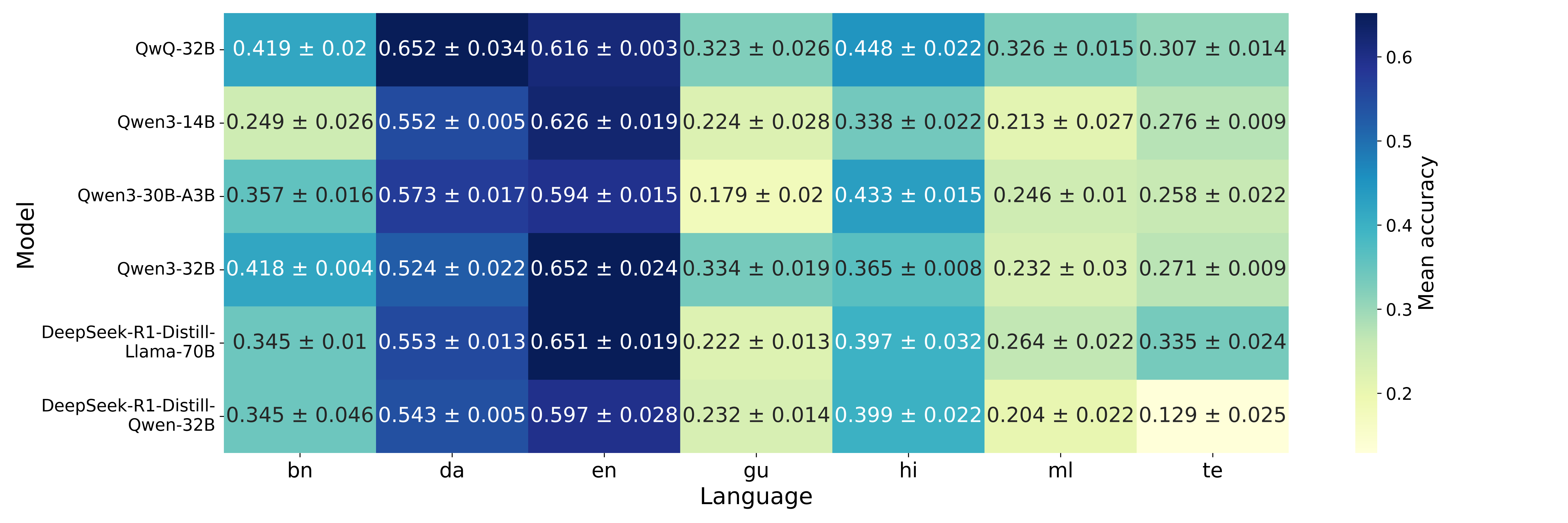}
    \caption{\textbf{GPQA diamond - Reasoning in the Question’s Language.} Final answer accuracy of LRMs on GPQA diamond when reasoning in question's language. Y-axis: models; X-axis: languages; each cell reports the avg@4 accuracy and standard deviation.
}
    \label{fig:gpqa_native_cot_reasoning}
\end{figure*}

\begin{figure*}[htbp]
    \centering
        \hfill
    \begin{subfigure}{0.48\textwidth} 
        \centering
        \includegraphics[width= 0.8\textwidth]{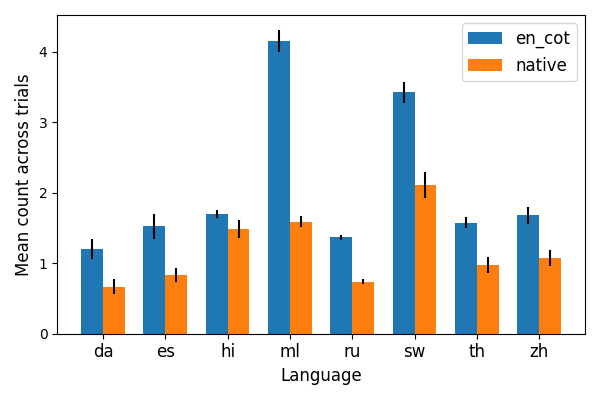} 
        \caption{Cognitive behavior: backtracking}
        \label{fig:cognitive_backtracking}
    
    \end{subfigure}
     \hfill
    \begin{subfigure}{0.48\textwidth} 
        \centering
        \includegraphics[width= 0.8\textwidth]{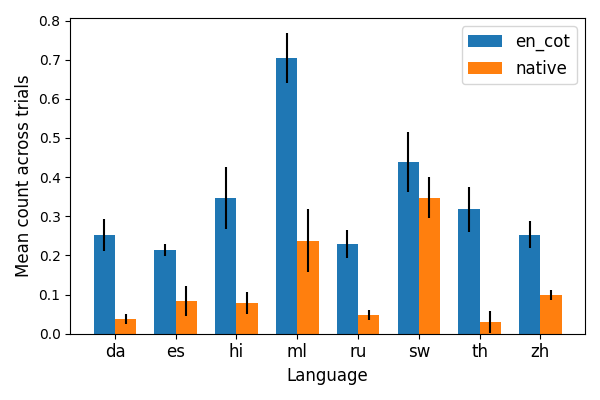} 
        \caption{Cognitive behavior: backward chaining }
        \label{fig:cognitive_backward_chaining}
    \end{subfigure}
    \hfill
    \begin{subfigure}{0.48\textwidth} 
        \centering
        \includegraphics[width= 0.8\textwidth]{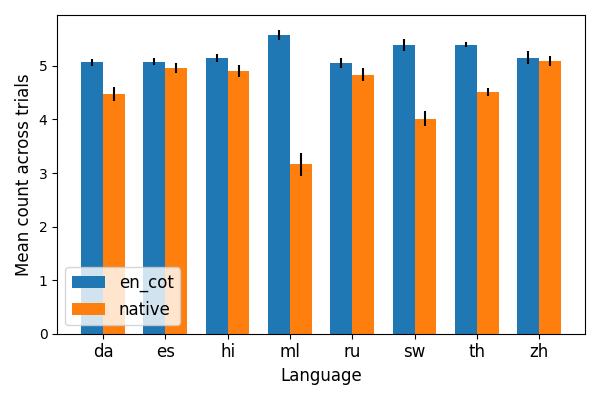} 
        \caption{Cognitive behavior: subgoal setting}
        \label{fig:cognitive_subgoal_setting}
    
    \end{subfigure}
    \begin{subfigure}{0.48\textwidth} 
        \centering
        \includegraphics[width= 0.8\textwidth]{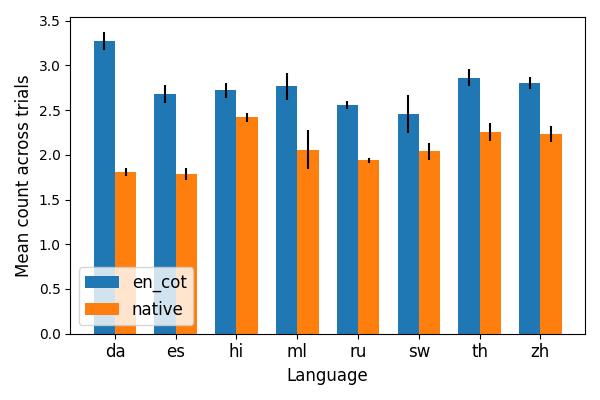} 
        \caption{Cognitive behavior: verification}
        \label{fig:cognitive_verification}
    
    \end{subfigure}
     \hfill
     \caption{\textbf{Cognitive behaviors in the reasoning chain} are averaged and compared for each language across reasoning in English and in the question's language. Error bars denote standard deviation. Model: Qwen QwQ-32B}
                
\label{fig:cognitive_behaviors}
\end{figure*}

\section{Results}
\label{others}
Figure~\ref{fig:native_cot_reasoning} presents the final answer accuracy of LRMs on the MGSM task when reasoning in the language of the question. We find that performance is highest for English, remains strong for high-resource languages, and declines progressively for lower-resource languages.

Figure~\ref{fig:Qwen3-32B MGSM} compares reasoning in English with reasoning in the question’s language for the MGSM task. As we move from high-resource to low-resource languages, the accuracy gap between English and native-language reasoning widens, indicating that reasoning in English generally yields higher final-answer accuracy.



We next examine whether this pattern holds for the GPQA diamond task which requires expert-level domain knowledge. In Figure~\ref{fig:gpqa_native_cot_reasoning} we observe that the contrast between performance in English and non-English languages is much higher than for the MGSM task during reasoning in the question's language. In addition, in Figure~\ref{fig:Qwen3-32B GPQA}, we could see the contrast between reasoning in English and question's language for multilingual gpqa questions is much larger than the same for MGSM (c.f. Figure~\ref{fig:en_cot_vs_native_cot}). Notably, Danish exhibits a smaller final-answer accuracy gap between reasoning in the question’s original language and reasoning in English (c.f. Figure~\ref{fig:Qwen3-32B GPQA}). Danish performs relatively well on GPQA-Diamond compared with other non-English languages (c.f. Figure~\ref{fig:gpqa_native_cot_reasoning}). This likely reflects Danish’s relatively high resource status among the languages analyzed (c.f. Section. \ref{sec: Grouping Languages by Resource Availability}). These results highlight that leveraging English is particularly important for tapping an LLM’s domain knowledge and reasoning ability on expert-level questions. Additional results are shown in appendix \ref{sec: Results: Additional details}.



\section{Analyzing cognitive behaviors in the reasoning trace}
\label{sec: Analyzing cognitive behaviors in the reasoning trace}

Evaluating multilingual reasoning by final answer accuracy alone overlooks key aspects of an LRM’s response generation. While final answer accuracy typically captures the essential behavior of LLMs, LRMs introduce additional complexity through their reasoning traces.

To capture this, following prior studies \citep{gandhi2025cognitive}, we analyze reasoning traces enclosed in <think></think> tokens for cognitive attributes such as \textit{Sub-goal setting} (breaking a task into smaller, manageable steps), \textit{Verification} (performing systematic checks
to catch and correct errors), \textit{Back-tracking} (abandoning unsuccessful strategies and selecting alternative approaches) , and \textit{Backward chaining} (reasoning from the desired outcome back toward the necessary initial inputs). These behaviors reflect how expert practitioners tackle difficult problems; for example, a software engineer tests and reviews each change, rewrites code when a design fails, and splits large features into smaller, testable modules. 

We use gpt-4o-mini as a judge to identify and count these attributes, employing evaluation prompts (c.f. Appendix \ref{sec: Analyzing Cognitive Behaviors: Additional details}) adapted from prior studies \citep{gandhi2025cognitive}. For the MGSM task, we compute the average frequency of each attribute for the Qwen QwQ-32B model across languages, and report means and standard deviations over four trials. Appendix \ref{sec: Analyzing Cognitive Behaviors: human lm agreement} details the human-LLM agreement study for this task.

Figure \ref{fig:cognitive_behaviors} illustrates the distribution of cognitive traits within the reasoning traces of an LRM. These traits appear more frequently when the model reasons in English than when it reasons in the question's language, and they coincide with higher final-answer accuracy. Among the attributes, sub-goal setting and verification appear more prominently in question's language reasoning trace, whereas back-tracking and backward chaining are comparatively less frequent.

\begin{figure*}[t]
    \centering
        \hfill
    \begin{subfigure}{0.48\textwidth} 
        \centering
        \includegraphics[width= \textwidth]{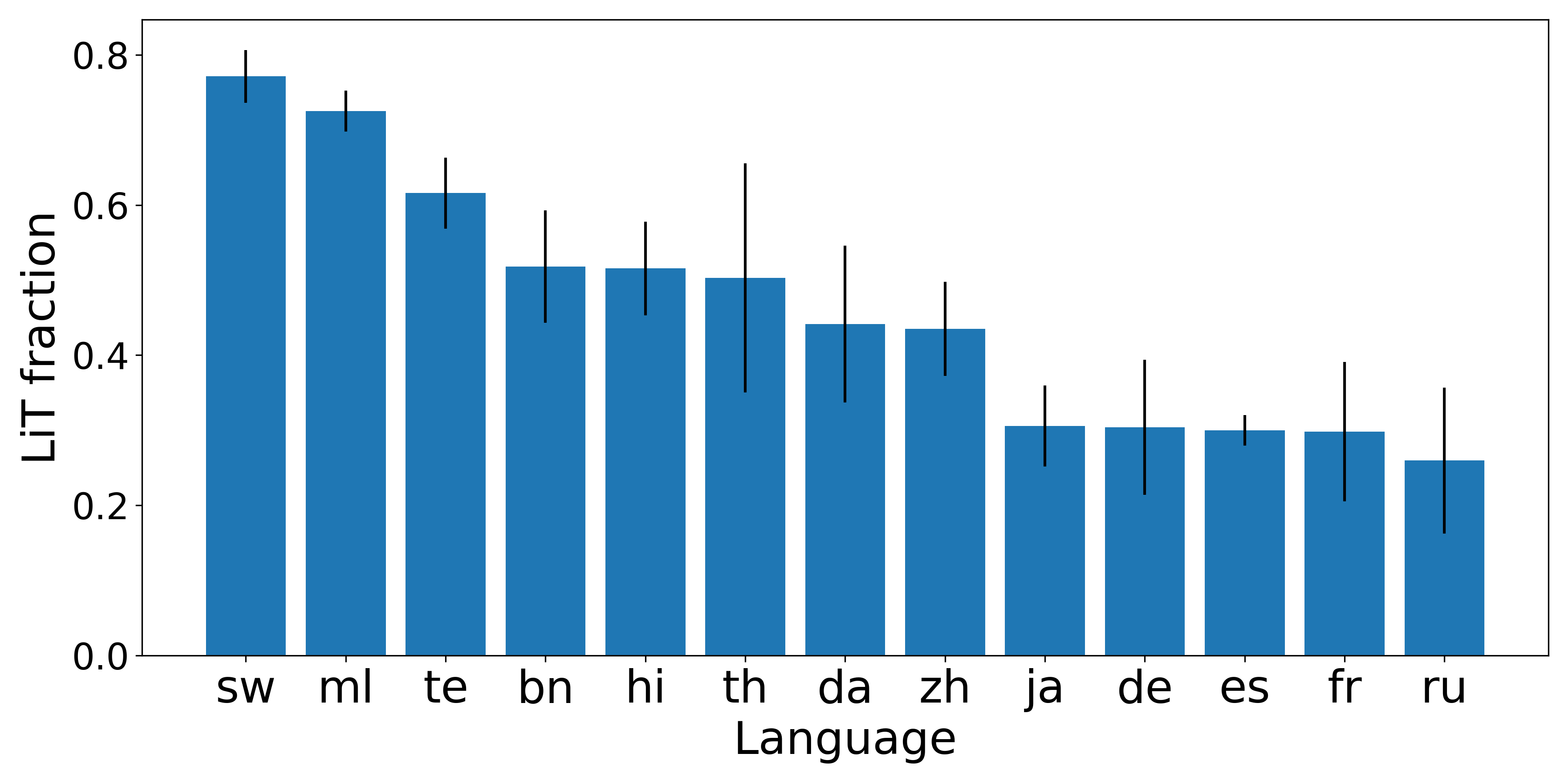} 
        \caption{MGSM}
        \label{fig:mgsm_lit}
    
    \end{subfigure}
     \hfill
    \begin{subfigure}{0.48\textwidth} 
        \centering
        \includegraphics[width= 0.85\textwidth]{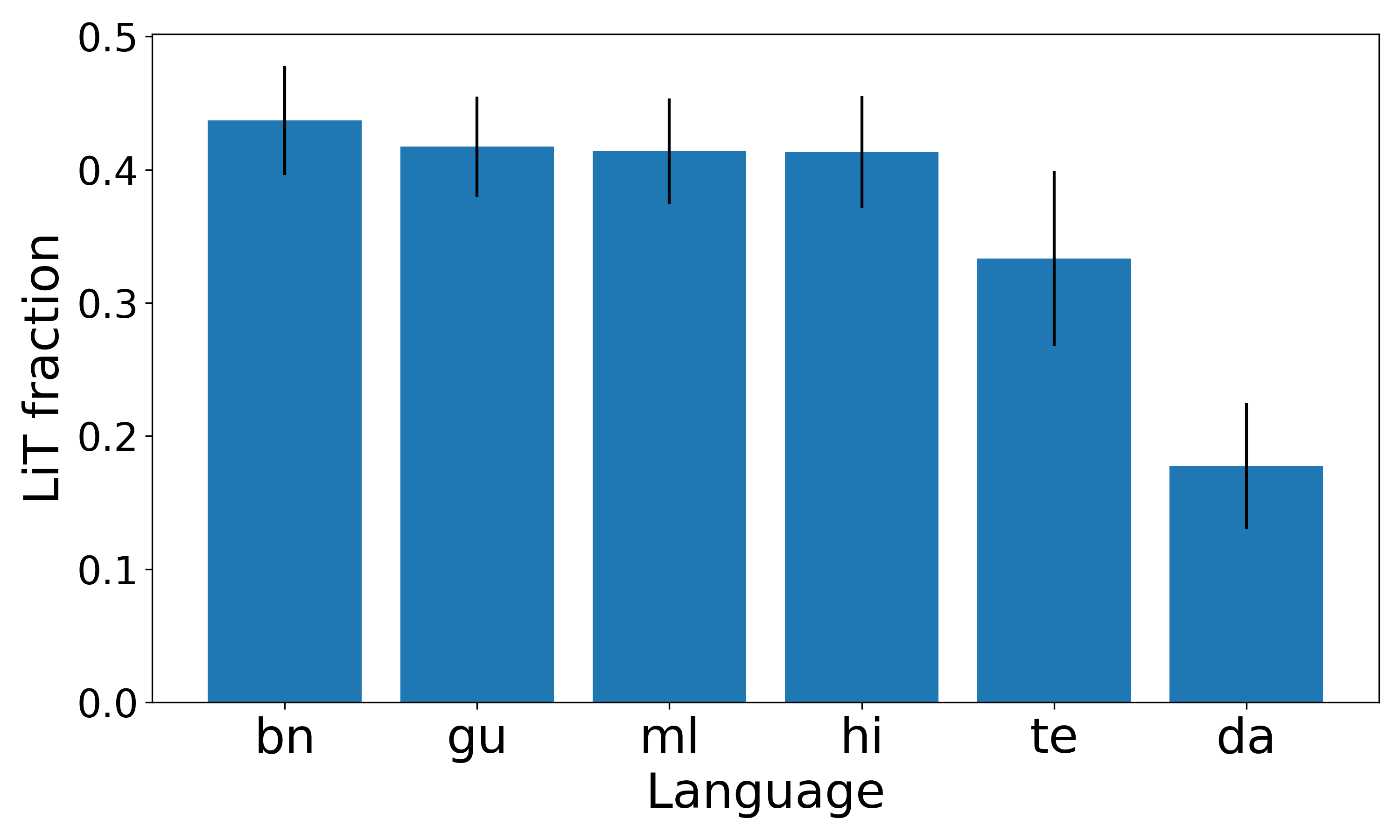} 
        \caption{GPQA }
        \label{fig:gpqa_lit}
    \end{subfigure}
    \hfill
     \caption{\textbf{Lost in Translation.} Visualization of fraction of incorrect answers occurring due to errors in translation when reasoning in English. Error bars represents standard deviation.}
                
\label{fig: lit}
\end{figure*}

\section{Making a case for reasoning in the Question's language}
\label{section: case for reasoning in native language}
In analyzing reasoning in English versus the question’s language, we generally observe stronger performance in English. Yet, a key question remains: \textit{are there cases where native-language reasoning surpasses English?} We find such cases, typically when translation introduces errors into the English reasoning. We term this phenomenon “Lost in Translation,” and provide a detailed explanation and example in Appendix \ref{sec: lost in translation}.

To quantify “Lost in Translation,” we measure the fraction of incorrect answers from English reasoning caused by translation mistakes; errors that would not arise if reasoning occurred directly in the question's language. These cases are identified using gpt-4o-mini as evaluator. The evaluation setup and prompts are described in Appendix \ref{sec: lost in translation quantitative analysis}. We average the proportion across four trials with the Qwen QwQ-32B model and report the standard deviation. This averaged proportion is referred to as the LiT (Lost in Translation) fraction. Appendix \ref{sec: lost in translation Human LM agreement}  details the human-LLM agreement study for the Lost in Translation (LiT) task.

Figure \ref{fig:mgsm_lit} reports the LiT fraction for MGSM, decreasing from 0.77 in low-resource languages to 0.30 in high-resource languages, showing that translation errors are more common in low-resource settings. Similarly, Figure \ref{fig:gpqa_lit} shows LiT fractions for GPQA, ranging from 0.44 to 0.33.

A substantial portion of incorrect answers can be traced to errors introduced during translation, exposing a systemic weakness in reasoning in English. While our prior experiments have shown that reasoning in English allows an LLM to better leverage its domain-specific knowledge and reasoning capabilities, this experiment reveals a critical vulnerability: \textbf{in multilingual tasks, English reasoning risks being “Lost in Translation”.}

\section{Conclusion}


Overall, we find that models generally reason better in English than in the language of the question. This advantage is not limited to final-answer accuracy, but also in the presence of cognitive behaviors in the reasoning chain. However, relying on English introduces a critical weakness: translation of multilingual inputs into English sometimes produces errors that would likely be avoided if models reasoned directly in the question's language. These findings underscore the need to develop and benchmark native-language reasoning capabilities with the same rigor and resources devoted to English.


We address an important question: are LRMs that primarily reason in English by translating non-English inputs sufficient for robust multilingual reasoning? Our results suggest they are not. Achieving reliable native-language reasoning will require targeted efforts in dataset construction, training objectives, and evaluation that preserve language-specific reasoning patterns. This study contributes an empirical diagnosis of the problem and establishes baselines that future work can use to measure progress toward native-language reasoning.



\section{Limitations}
Our evaluation of multilingual reasoning is confined to mathematical and scientific tasks; whether the patterns we observe extend to open-ended questions or other task types remains an open question. Likewise, our experiments are limited to open-weight LRMs, so it remains to be seen if these findings generalize to closed/proprietary models such as o1 \citep{jaech2024openai} and Gemini 2.5 Pro \citep{comanici2025gemini}. Future work should evaluate these axes to determine the broader applicability of our results.

\section{Ethics Statement}
Through this work, our aim is to analyze the performance of LRMs when reasoning in English vs reasoning in questions language for multilingual tasks. We emphasize that it is not our intention to diminish the value or significance  of non-English languages included in this study.

The code and datasets created in this work will be made available under permissible licenses. Generative AI systems were only used for assistance purely with the language of the paper, e.g., paraphrasing, spell-check, polishing the author’s original content, and for writing boiler-plate code.

\section{Acknowledgments}
We thank AI4Bharat lab members Deepon Halder, Irina Agastin, and Atharva Gawane for their assistance with annotation on the cognitive behavior analysis and Lost in Translation task. We would also like to thank EkStep Foundation and
Nilekani Philanthropies for their generous grant
towards research at AI4Bharat.

\bibliography{custom}

\begin{thebibliography}{23}
\providecommand{\natexlab}[1]{#1}

\bibitem[{Aggarwal et~al.(2025)Aggarwal, Tanmay, Agrawal, Ayush, Palangi, and Liang}]{aggarwal2025language}
Tushar Aggarwal, Kumar Tanmay, Ayush Agrawal, Kumar Ayush, Hamid Palangi, and Paul~Pu Liang. 2025.
\newblock Language models' factuality depends on the language of inquiry.
\newblock \emph{arXiv preprint arXiv:2502.17955}.

\bibitem[{Anonymous(2025)}]{our2023}
Anonymous. 2025.
\newblock Title omitted for double-blind review.
\newblock In \emph{Under review}.
\newblock Anonymized for review.

\bibitem[{Comanici et~al.(2025)Comanici, Bieber, Schaekermann, Pasupat, Sachdeva, Dhillon, Blistein, Ram, Zhang, Rosen et~al.}]{comanici2025gemini}
Gheorghe Comanici, Eric Bieber, Mike Schaekermann, Ice Pasupat, Noveen Sachdeva, Inderjit Dhillon, Marcel Blistein, Ori Ram, Dan Zhang, Evan Rosen, and 1 others. 2025.
\newblock Gemini 2.5: Pushing the frontier with advanced reasoning, multimodality, long context, and next generation agentic capabilities.
\newblock \emph{arXiv preprint arXiv:2507.06261}.

\bibitem[{Deepinfra(2025)}]{DeepInfraAPI2025}
Deepinfra. 2025.
\newblock Deepinfra api.
\newblock \url{https://deepinfra.com/docs/api-reference}.
\newblock Accessed 2024 - 09-21.

\bibitem[{Gandhi et~al.(2025)Gandhi, Chakravarthy, Singh, Lile, and Goodman}]{gandhi2025cognitive}
Kanishk Gandhi, Ayush Chakravarthy, Anikait Singh, Nathan Lile, and Noah~D Goodman. 2025.
\newblock Cognitive behaviors that enable self-improving reasoners, or, four habits of highly effective stars.
\newblock \emph{arXiv preprint arXiv:2503.01307}.

\bibitem[{Gao et~al.(2024)Gao, Tow, Abbasi, Biderman, Black, DiPofi, Foster, Golding, Hsu, Le~Noac'h, Li, McDonell, Muennighoff, Ociepa, Phang, Reynolds, Schoelkopf, Skowron, Sutawika, Tang, Thite, Wang, Wang, and Zou}]{eval-harness}
Leo Gao, Jonathan Tow, Baber Abbasi, Stella Biderman, Sid Black, Anthony DiPofi, Charles Foster, Laurence Golding, Jeffrey Hsu, Alain Le~Noac'h, Haonan Li, Kyle McDonell, Niklas Muennighoff, Chris Ociepa, Jason Phang, Laria Reynolds, Hailey Schoelkopf, Aviya Skowron, Lintang Sutawika, and 5 others. 2024.
\newblock \href {https://doi.org/10.5281/zenodo.12608602} {The language model evaluation harness}.

\bibitem[{Guo et~al.(2025)Guo, Yang, Zhang, Song, Zhang, Xu, Zhu, Ma, Wang, Bi et~al.}]{guo2025deepseek}
Daya Guo, Dejian Yang, Haowei Zhang, Junxiao Song, Ruoyu Zhang, Runxin Xu, Qihao Zhu, Shirong Ma, Peiyi Wang, Xiao Bi, and 1 others. 2025.
\newblock Deepseek-r1: Incentivizing reasoning capability in llms via reinforcement learning.
\newblock \emph{arXiv preprint arXiv:2501.12948}.

\bibitem[{Jaech et~al.(2024)Jaech, Kalai, Lerer, Richardson, El-Kishky, Low, Helyar, Madry, Beutel, Carney et~al.}]{jaech2024openai}
Aaron Jaech, Adam Kalai, Adam Lerer, Adam Richardson, Ahmed El-Kishky, Aiden Low, Alec Helyar, Aleksander Madry, Alex Beutel, Alex Carney, and 1 others. 2024.
\newblock Openai o1 system card.
\newblock \emph{arXiv preprint arXiv:2412.16720}.

\bibitem[{{OpenAI}(2024)}]{openai_gpt4o_mini_2024}
{OpenAI}. 2024.
\newblock \href {https://openai.com/index/gpt-4o-mini-advancing-cost-efficient-intelligence/} {{GPT-4o mini: advancing cost-efficient intelligence}}.
\newblock Blog post.
\newblock Accessed: 2025-8-06.

\bibitem[{Qi et~al.(2025)Qi, Chen, Xiong, Fern{\'a}ndez, Bitterman, and Bisazza}]{qi2025models}
Jirui Qi, Shan Chen, Zidi Xiong, Raquel Fern{\'a}ndez, Danielle~S Bitterman, and Arianna Bisazza. 2025.
\newblock When models reason in your language: Controlling thinking trace language comes at the cost of accuracy.
\newblock \emph{arXiv preprint arXiv:2505.22888}.

\bibitem[{{QwenTeam}(2025)}]{qwen_qwq_32b}
{QwenTeam}. 2025.
\newblock Qwen qwq-32b.
\newblock \url{https://qwen.ai/blog?id=6aed6aa257238a0b6c77a6753f180350c2fecc4a&from=research.research-list}.
\newblock Accessed: 2025-8-06.

\bibitem[{Rein et~al.(2024)Rein, Hou, Stickland, Petty, Pang, Dirani, Michael, and Bowman}]{rein2024gpqa}
David Rein, Betty~Li Hou, Asa~Cooper Stickland, Jackson Petty, Richard~Yuanzhe Pang, Julien Dirani, Julian Michael, and Samuel~R Bowman. 2024.
\newblock Gpqa: A graduate-level google-proof q\&a benchmark.
\newblock In \emph{First Conference on Language Modeling}.

\bibitem[{SarvamAI(2025)}]{sarvam_translate}
SarvamAI. 2025.
\newblock Sarvam translate.
\newblock \url{https://www.sarvam.ai/blogs/sarvam-translate}.
\newblock Accessed: 2025-7-06.

\bibitem[{Shi et~al.(2022)Shi, Suzgun, Freitag, Wang, Srivats, Vosoughi, Chung, Tay, Ruder, Zhou et~al.}]{shi2022language}
Freda Shi, Mirac Suzgun, Markus Freitag, Xuezhi Wang, Suraj Srivats, Soroush Vosoughi, Hyung~Won Chung, Yi~Tay, Sebastian Ruder, Denny Zhou, and 1 others. 2022.
\newblock Language models are multilingual chain-of-thought reasoners.
\newblock \emph{arXiv preprint arXiv:2210.03057}.

\bibitem[{Tam et~al.(2025)Tam, Wu, Chiu, Lin, Chen, and Lee}]{languagemattersappier}
Zhi~Rui Tam, Cheng{-}Kuang Wu, Yu~Ying Chiu, Chieh{-}Yen Lin, Yun{-}Nung Chen, and Hung{-}yi Lee. 2025.
\newblock \href {https://doi.org/10.48550/arXiv.2505.17407} {Language matters: How do multilingual input and reasoning paths affect large reasoning models?}
\newblock abs/2505.17407.

\bibitem[{Team et~al.(2024)Team, Riviere, Pathak, Sessa, Hardin, Bhupatiraju, Hussenot, Mesnard, Shahriari, Ram{\'e} et~al.}]{team2024gemma}
Gemma Team, Morgane Riviere, Shreya Pathak, Pier~Giuseppe Sessa, Cassidy Hardin, Surya Bhupatiraju, L{\'e}onard Hussenot, Thomas Mesnard, Bobak Shahriari, Alexandre Ram{\'e}, and 1 others. 2024.
\newblock Gemma 2: Improving open language models at a practical size.
\newblock \emph{arXiv preprint arXiv:2408.00118}.

\bibitem[{Touvron et~al.(2023)Touvron, Lavril, Izacard, Martinet, Lachaux, Lacroix, Rozi{\`e}re, Goyal, Hambro, Azhar et~al.}]{touvron2023llama}
Hugo Touvron, Thibaut Lavril, Gautier Izacard, Xavier Martinet, Marie-Anne Lachaux, Timoth{\'e}e Lacroix, Baptiste Rozi{\`e}re, Naman Goyal, Eric Hambro, Faisal Azhar, and 1 others. 2023.
\newblock Llama: Open and efficient foundation language models.
\newblock \emph{arXiv preprint arXiv:2302.13971}.

\bibitem[{Wang et~al.(2025)Wang, Zhang, Tang, Wei, Yang, Wang, Sun, Sun, Zhang, Wu et~al.}]{wang2025polymath}
Yiming Wang, Pei Zhang, Jialong Tang, Haoran Wei, Baosong Yang, Rui Wang, Chenshu Sun, Feitong Sun, Jiran Zhang, Junxuan Wu, and 1 others. 2025.
\newblock Polymath: Evaluating mathematical reasoning in multilingual contexts.
\newblock \emph{arXiv preprint arXiv:2504.18428}.

\bibitem[{Wendler et~al.(2024)Wendler, Veselovsky, Monea, and West}]{wendler_llama}
Chris Wendler, Veniamin Veselovsky, Giovanni Monea, and Robert West. 2024.
\newblock \href {https://doi.org/10.18653/v1/2024.acl-long.820} {Do llamas work in english? on the latent language of multilingual transformers}.
\newblock In \emph{Proceedings of the 62nd Annual Meeting of the Association for Computational Linguistics (Volume 1: Long Papers), {ACL} 2024, Bangkok, Thailand, August 11-16, 2024}, pages 15366--15394. Association for Computational Linguistics.

\bibitem[{Wu et~al.(2025)Wu, Yu, Yogatama, Lu, and Kim}]{semantic_hub}
Zhaofeng Wu, Xinyan~Velocity Yu, Dani Yogatama, Jiasen Lu, and Yoon Kim. 2025.
\newblock \href {https://openreview.net/forum?id=FrFQpAgnGE} {The semantic hub hypothesis: Language models share semantic representations across languages and modalities}.
\newblock In \emph{The Thirteenth International Conference on Learning Representations, {ICLR} 2025, Singapore, April 24-28, 2025}. OpenReview.net.

\bibitem[{Yang et~al.(2025)Yang, Li, Yang, Zhang, Hui, Zheng, Yu, Gao, Huang, Lv et~al.}]{yang2025qwen3}
An~Yang, Anfeng Li, Baosong Yang, Beichen Zhang, Binyuan Hui, Bo~Zheng, Bowen Yu, Chang Gao, Chengen Huang, Chenxu Lv, and 1 others. 2025.
\newblock Qwen3 technical report.
\newblock \emph{arXiv preprint arXiv:2505.09388}.

\bibitem[{Yong et~al.(2025)Yong, Adilazuarda, Mansurov, Zhang, Muennighoff, Eickhoff, Winata, Kreutzer, Bach, and Aji}]{yong2025crosslingual}
Zheng-Xin Yong, M~Farid Adilazuarda, Jonibek Mansurov, Ruochen Zhang, Niklas Muennighoff, Carsten Eickhoff, Genta~Indra Winata, Julia Kreutzer, Stephen~H Bach, and Alham~Fikri Aji. 2025.
\newblock Crosslingual reasoning through test-time scaling.
\newblock \emph{arXiv preprint arXiv:2505.05408}.

\bibitem[{Zhao et~al.(2024)Zhao, Zhang, Chen, Kawaguchi, and Bing}]{zhao_handle_multi}
Yiran Zhao, Wenxuan Zhang, Guizhen Chen, Kenji Kawaguchi, and Lidong Bing. 2024.
\newblock \href {http://papers.nips.cc/paper\_files/paper/2024/hash/1bd359b32ab8b2a6bbafa1ed2856cf40-Abstract-Conference.html} {How do large language models handle multilingualism?}
\newblock In \emph{Advances in Neural Information Processing Systems 38: Annual Conference on Neural Information Processing Systems 2024, NeurIPS 2024, Vancouver, BC, Canada, December 10 - 15, 2024}.

\end{thebibliography}

\appendix

\section{Related Work - additional details}
\label{sec: Related Work - additional details}
Prior work has shown that English-centric LLMs carry out intermediate reasoning in a language-agnostic latent space that is nevertheless biased toward English \citep{wendler_llama,zhao_handle_multi}. \citet{semantic_hub} further demonstrates that this English bias extends across modalities.

LRMs amplify this tendency by favoring explicit reasoning in English. In this paper we present a careful, empirical comparison of reasoning performed in English versus reasoning in the question’s original language, and we highlight a seldom-discussed shortcoming of English-centric reasoning; namely, that translation-related errors can undermine correctness even when English reasoning appears stronger overall.

\begin{figure*}[htbp]
  \centering
  \includegraphics[width= \textwidth]{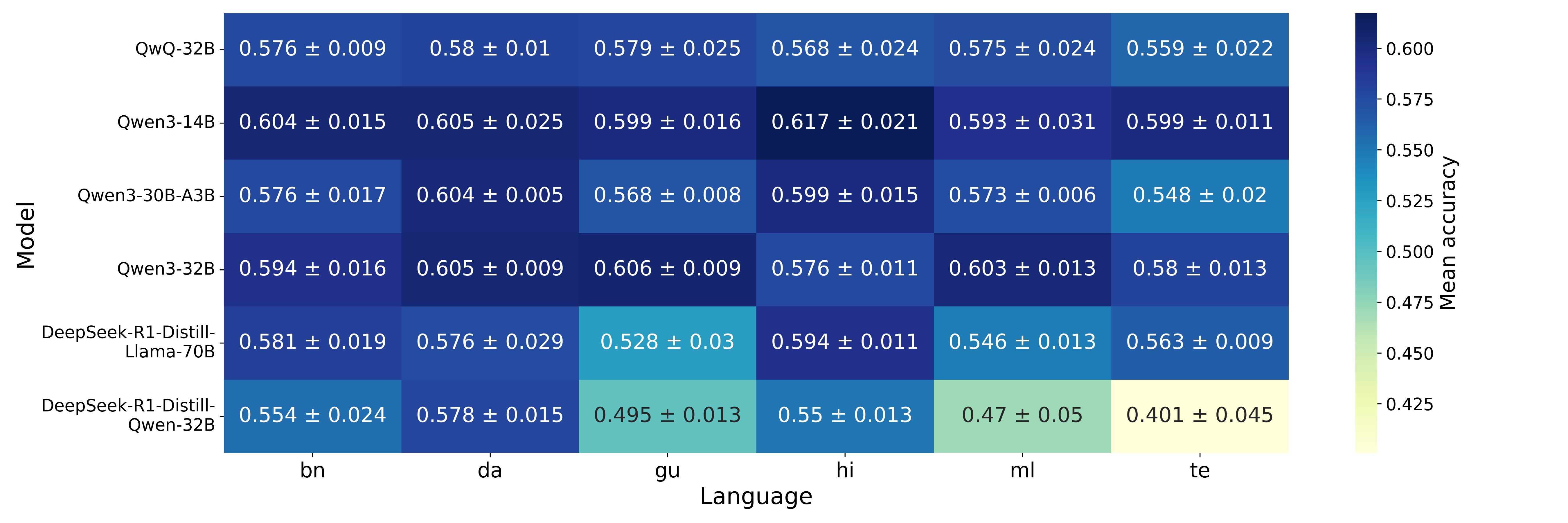}
    \caption{Here we compare performance of reasoning models on GPQA diamond task. Y axis represent LLMs, X axis represents languages and each entry is avg@4 accuracy score. Here the models are explicitly instructed to reason in English via prefix tokens. 
}
    \label{fig:gpqa_en_cot_reasoning}
\end{figure*}

\begin{figure*}[t]
  \centering
  \includegraphics[width=\textwidth]{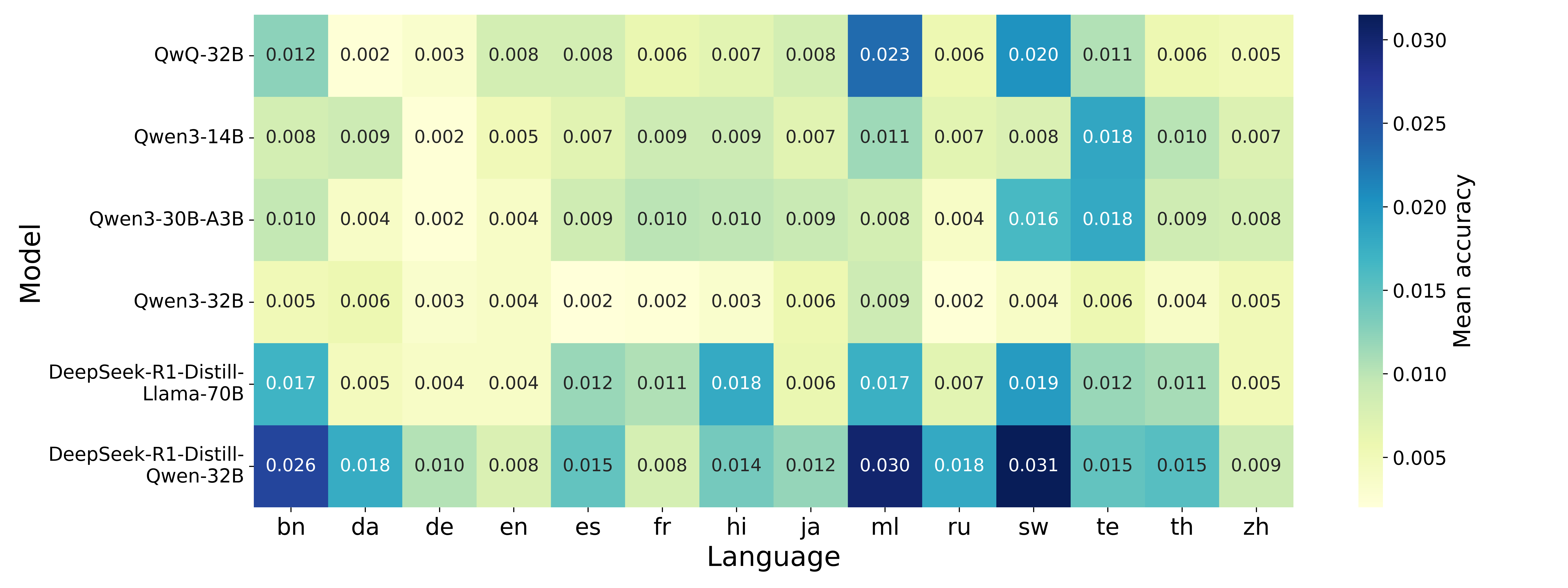}
    \caption{\textbf{Standard deviation} of accuracy of reasoning models on MGSM task. Y axis represent LLMs, X axis represents languages. Here the models are explicitly instructed to reason in question's language via prefix tokens.}
    \label{fig:native_cot_reasoning_std}
\end{figure*}


\section{Experimental settings: Additional details}
\label{sec: Experimental setting additional details}
Our evaluation framework is based on lm-evaluation-harness library \citep{eval-harness}. We update the framework to make it suitable for evaluating reasoning models. This includes capturing the final answer in a \verb|\boxed{}| format and extracting the answer from there. 

We set the maximum output token limit to 10,000 for MGSM and 30,000 for GPQA diamond dataset. These limits were chosen based on pilot runs. GPQA-Diamond required a larger limit because the higher task difficulty here led to substantially longer reasoning traces; MGSM’s responses were much shorter on average.

Because we used sampling-based decoding rather than greedy decoding, we ran the experiment four times and report the mean to ensure reproducibility.

\subsection{Dataset}
\label{sec:dataset}
The MGSM dataset \citep{shi2022language} involves 250 mathematical question-answer pairs at graduate-level for 12 languages. We extend this dataset to Hindi and Malayalam using Sarvam - Translate \citep{sarvam_translate} and to Danish using \citet{comanici2025gemini}. The quality of these translations are manually verified to ensure high quality. \\

GPQA Diamond: The GPQA Diamond dataset \citep{rein2024gpqa} contains 198 challenging science questions that require advanced subject expertise. It is significantly harder than the MGSM dataset. In this study, we use an in-house translation of GPQA Diamond into five Indic languages: Hindi, Gujarati, Bengali, Malayalam, and Telugu. Additionally, we translate GPQA Diamond dataset into Danish using \citet{comanici2025gemini}.

\subsection{Models}
\label{sec: models}
In this study, we consider open weight reasoning models including Qwen QwQ-32B \citep{qwen_qwq_32b}, Qwen3-32B \citep{yang2025qwen3}, Qwen3-14B \citep{yang2025qwen3}, Qwen3-30-A3B \citep{yang2025qwen3}, DeepSeek-R1-Distill-Llama-70B \citep{guo2025deepseek}, DeepSeek-R1-Distill-Qwen-32B \citep{yang2025qwen3}.

\subsection{Languages Involved}
\label{sec:Language Involved}

MGSM involves a typologically diverse set of ten languages other than English (en),
spanning eight language families. They are Bengali (bn), Chinese (zh),
French (fr), German (de), Japanese (ja), Russian (ru),
Spanish (es), Swahili (sw), Telugu (te), and Thai (th).
We extended MGSM to Malayalam (ml), Hindi (hi) and Danish (da). 
GPQA diamond is originally in English and is translated to Danish (da) and 5 Indic languages: Hindi (hi), Malayalam (ml), Gujarati (Gu), Bengali (bn) and Telugu (te).

Most of the European and South Asian languages listed belong to the Indo-European family: French and Spanish are Romance languages, German is West Germanic, Danish is North Germanic, Russian is East Slavic, and Bengali, Hindi and Gujarati are Indo-Aryan languages. The remaining languages come from other families: Malayalam and Telugu are Dravidian, Chinese is Sino-Tibetan, Thai is Kra-Dai, Japanese is Japonic, and Swahili is Niger–Congo.

\subsection{API Cost Overview}
 Qwen QwQ-32B \citep{qwen_qwq_32b}, Qwen3-32B \citep{yang2025qwen3}, Qwen3-14B \citep{yang2025qwen3}, Qwen3-30-A3B \citep{yang2025qwen3}, DeepSeek-R1-Distill-Llama-70B \citep{guo2025deepseek}, DeepSeek-R1-Distill-Qwen-32B \citep{yang2025qwen3}.   was accessed via deepinfra API \citep{DeepInfraAPI2025}. gpt-4o-mini \citep{openai_gpt4o_mini_2024} was accessed via openai API. Table \ref{table:api-costs} provides detailed cost incurred for each model. As inferred the total cost incurred via API usage for this project was 295 dollars.

\begin{table*}[htbp]
  \centering
  \begin{threeparttable}
  \begin{tabularx}{0.75\textwidth}{@{} X l >{\centering\arraybackslash}p{2.0cm} @{}}
    \toprule
    \textbf{Model} & \textbf{Provider} &  \textbf{Cost (USD)} \\
    \midrule
    Qwen QwQ-32B                  & Deepinfra &  37\\
    Qwen3-32B                     & Deepinfra & 29 \\
    Qwen3-14B                     & Deepinfra & 26 \\
    Qwen3-30-A3B                  & Deepinfra & 28 \\
    DeepSeek-R1-Distill-Llama-70B & Deepinfra &  42 \\
    DeepSeek-R1-Distill-Qwen-32B  & Deepinfra & 29 \\
    gpt-4o-mini                   & OpenAI & 104 \\
    \midrule

    \multicolumn{2}{l}{\textbf{Grand Total (USD)}} & \textbf{295} \\
    \bottomrule
  \end{tabularx}
  \caption{API usage and cost for models utilized in this study.}
   \label{table:api-costs}
  \end{threeparttable}
\end{table*}

\subsection{Prepending Prefix tokens}
\label{sec: Prepending Prefix tokens}
The LLM’s reasoning is enclosed within <think></think> tokens. Immediately after the <think> token, we insert a prefix token (a translation of the phrase “Ok, let me think step by step”) to force reasoning in a desired language. Figure~\ref{fig:think-prefix-language} illustrates the addition of the prefix token.

\begin{table*}[tb]
    \centering
    \begin{tabular}{|l|p{0.6\textwidth}|}
        \hline
        \textbf{Scenario} & \textbf{Token Sequence} \\
        \hline
        Without Language Forcing & \texttt{<|im\_start|>assistant\textbackslash n<think>\textbackslash n} \\
        \hline
        With Language Forcing & \texttt{<|im\_start|>assistant\textbackslash n<think>\textbackslash nD'accord, laissez-moi essayer de résoudre ce problème étape par étape.} \\
        \hline
    \end{tabular}
    \caption{Illustration of language forcing to control the language of thought. By inserting a specific language prefix (e.g., French) after the \texttt{<think>} token, the model is prompted to generate its internal reasoning in that language.}
    \label{fig:think-prefix-language}
\end{table*}

\subsection{System prompt}
\label{sec: sys prompt details}
Table \ref{tab:system_prompts_centered} showcases system prompts used for MGSM and GPQA Diamond tasks.

\begin{table*}[h!]
\centering

\renewcommand{\arraystretch}{1.8} 

\begin{tabularx}{0.95\textwidth}{@{}llX@{}}
\toprule
\textbf{Task} & \textbf{Reasoning Language} & \textbf{System Prompt} \\
\midrule

\multirowcell{2}{MGSM} &
  question's language &
  "You are a helpful assistant. You should reason and analyze the question in the language of the question and wrap your thought process in \texttt{<think>...</think>} tags. Then provide the answer in the language of the question, and keep the final answer in \textbackslash boxed\{\}." \\
  \cmidrule(l){2-3}
&
  English &
  "You are a helpful assistant. You should reason and analyze the question in English and wrap your thought process in \texttt{<think>...</think>} tags. Then provide the answer in the language of the question, and keep the final answer in \textbackslash boxed\{\}." \\
\midrule

\multirowcell{2}{GPQA Diamond} &
  question's language &
  "You are a helpful assistant. You should reason and analyze the question in the language of the question and wrap your thought process in \texttt{<think>...</think>} tags. Outside the \texttt{<think>...</think>} tags summarize the answer in the language of the question. Finally, select the correct option out of the given options A,B,C or D and keep this option in \textbackslash boxed\{\}." \\
  \cmidrule(l){2-3}
&
  English &
  "You are a helpful assistant. You should reason and analyze the question in English and wrap your thought process in \texttt{<think>...</think>} tags. Outside the \texttt{<think>...</think>} tags summarize the answer in the language of the question. Finally, select the correct option out of the given options A,B,C or D and keep this option in \textbackslash boxed\{\}." \\
\bottomrule
\end{tabularx}
\caption{System Prompt Variations for MGSM and GPQA Benchmarks}
\label{tab:system_prompts_centered}
\end{table*}

\section{Results: Additional details}
\label{sec: Results: Additional details}
 In Figure~\ref{fig:gpqa_native_cot_reasoning} we observe a substantial contrast between performance in English and non-English languages when reasoning is conducted in the question’s  language. However, as we go to Figure \ref{fig:gpqa_en_cot_reasoning} where the model reasons in English irrespective of the language of the questions, LLM's performance across languages seems more comparable. 

\subsection{English vs Question's language reasoning - Additional Models}
\label{sec: English vs Question's language reasoning - Additional Models}

\begin{figure*}[htbp]
    \centering
        \hfill
    \begin{subfigure}{0.48\textwidth} 
        \centering
        \includegraphics[width= \textwidth]{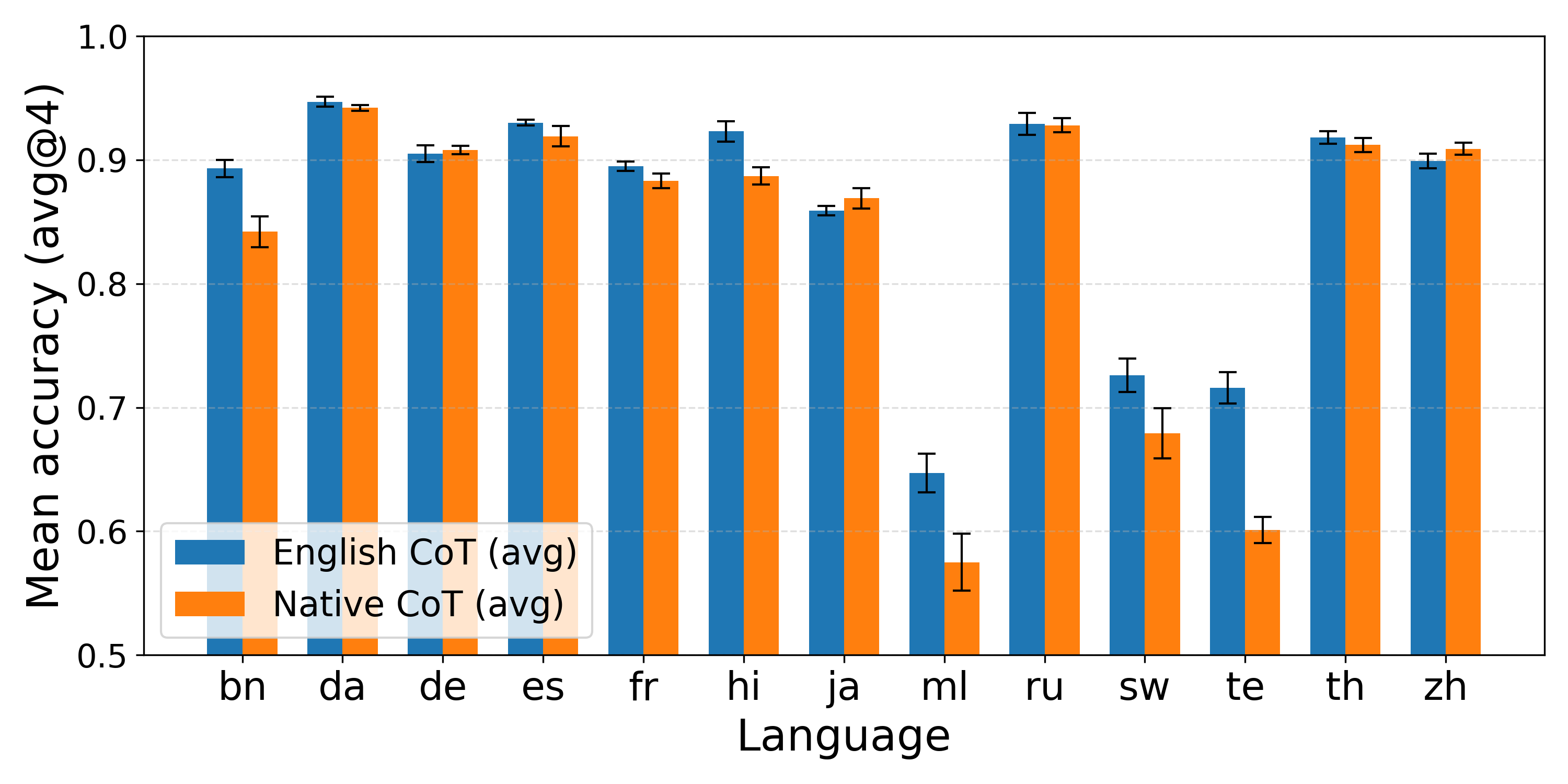} 
        \caption{MGSM}
        \label{fig:Qwen__QwQ-32B MGSM}
    
    \end{subfigure}
     \hfill
    \begin{subfigure}{0.48\textwidth} 
        \centering
        \includegraphics[width= \textwidth]{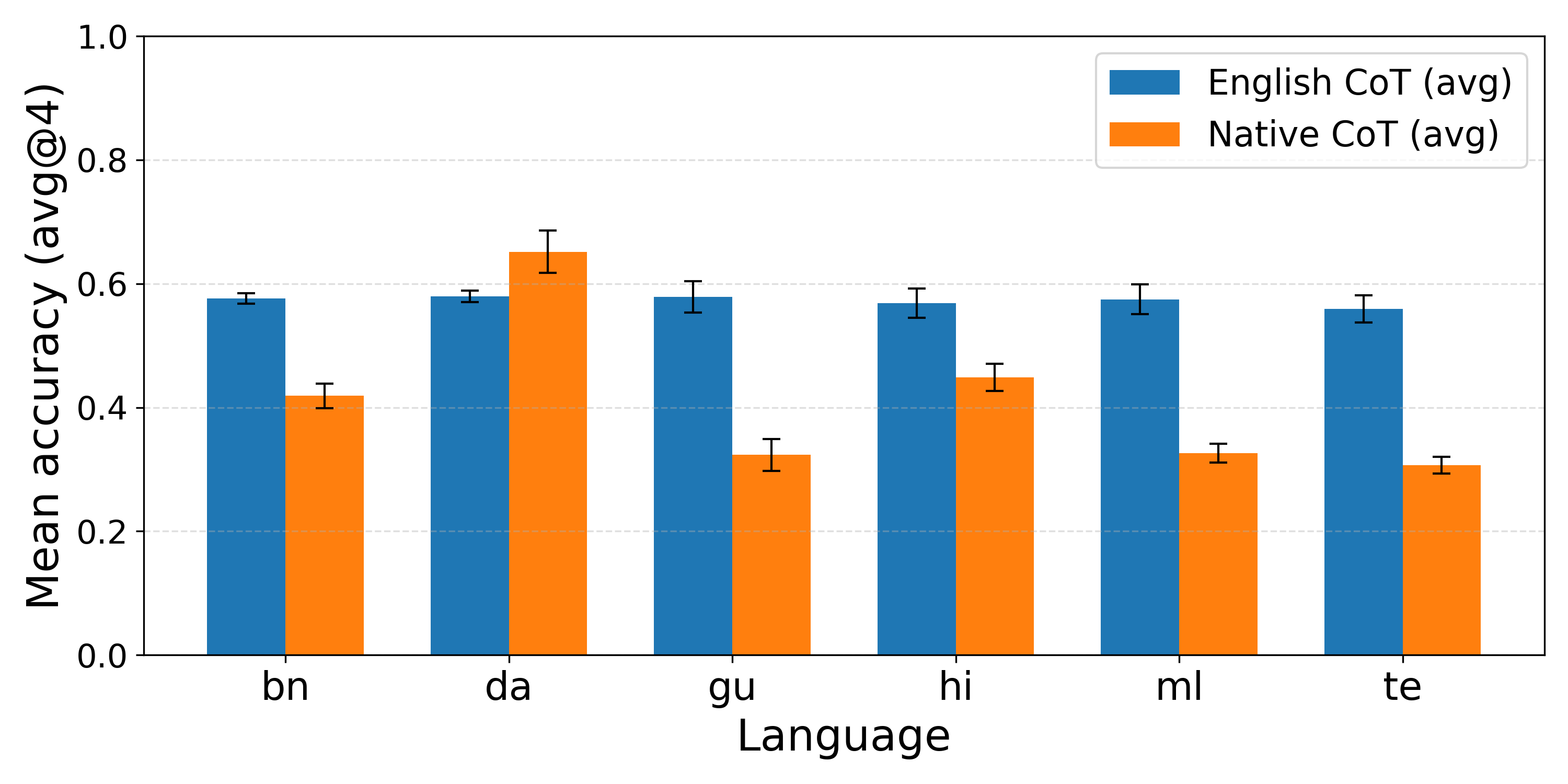} 
        \caption{GPQA Diamond}
        \label{fig:Qwen__QwQ-32B GPQA}
    \end{subfigure}
     \hfill
     \caption{\textbf{Reasoning in English vs Question's Language} is compared via final answer accuracy for Qwen QwQ 32B for MGSM and GPQA diamond task. Error bars denote standard deviation.}
                
\label{fig:en_cot_vs_native_cot qwen QwQ 32b}
\end{figure*}

\begin{figure*}[h!]
    \centering
        \hfill
    \begin{subfigure}{0.48\textwidth} 
        \centering
        \includegraphics[width= \textwidth]{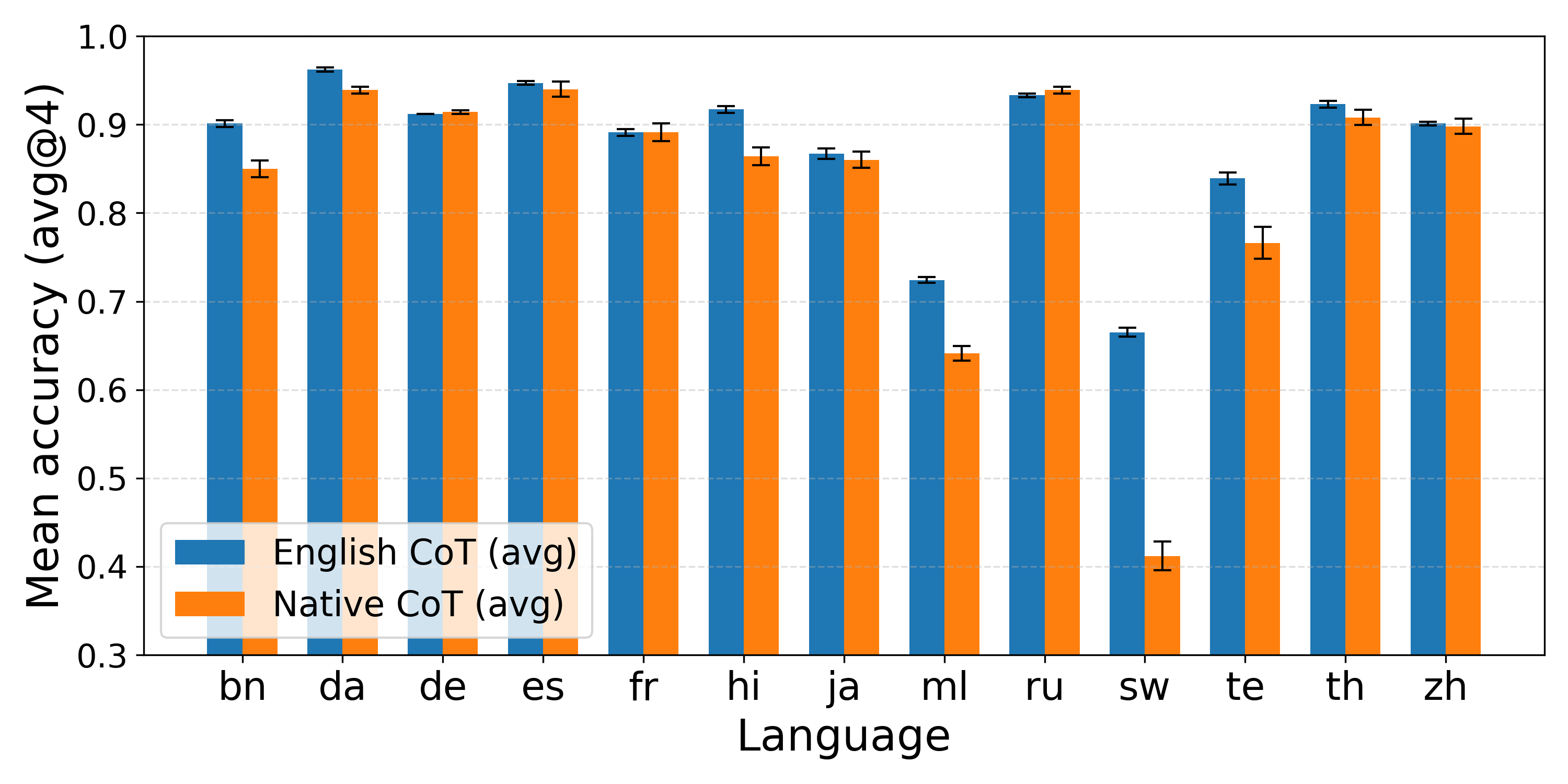} 
        \caption{MGSM}
        \label{fig:Qwen_Qwen3-30B-A3B MGSM}
    
    \end{subfigure}
     \hfill
    \begin{subfigure}{0.48\textwidth} 
        \centering
        \includegraphics[width= \textwidth]{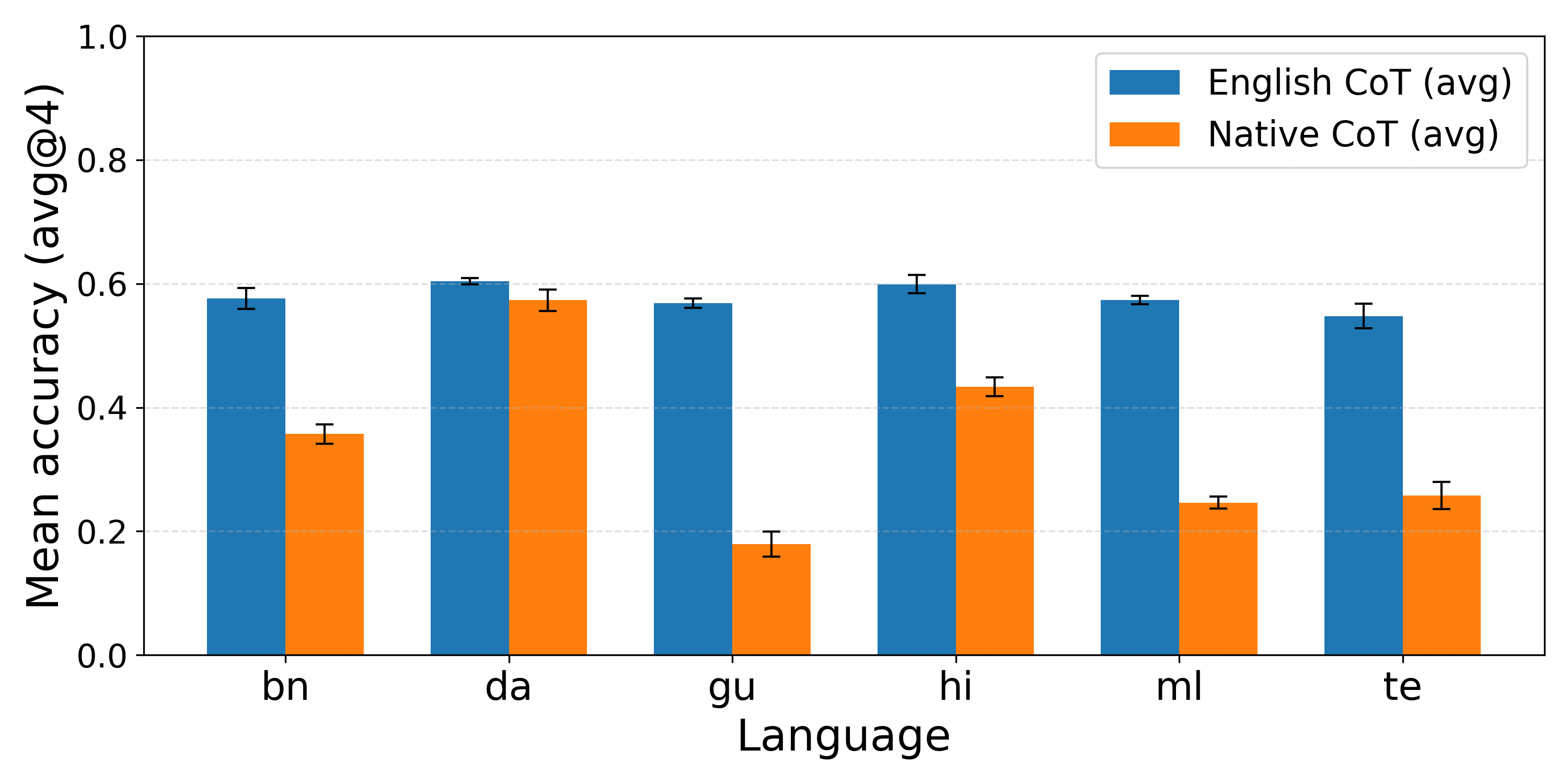} 
        \caption{GPQA Diamond}
        \label{fig:Qwen_Qwen3-30B-A3B GPQA}
    \end{subfigure}
     \hfill
     \caption{\textbf{Reasoning in English vs Question's Language} is compared via final answer accuracy for \textbf{Qwen3 30B-A3B} for MGSM and GPQA diamond task. Error bars denote standard deviation.}
                
\label{fig:en_cot_vs_native_cot Qwen3-30B-A3B}
\end{figure*}

\begin{figure*}[h!]
    \centering
        \hfill
    \begin{subfigure}{0.48\textwidth} 
        \centering
        \includegraphics[width= \textwidth]{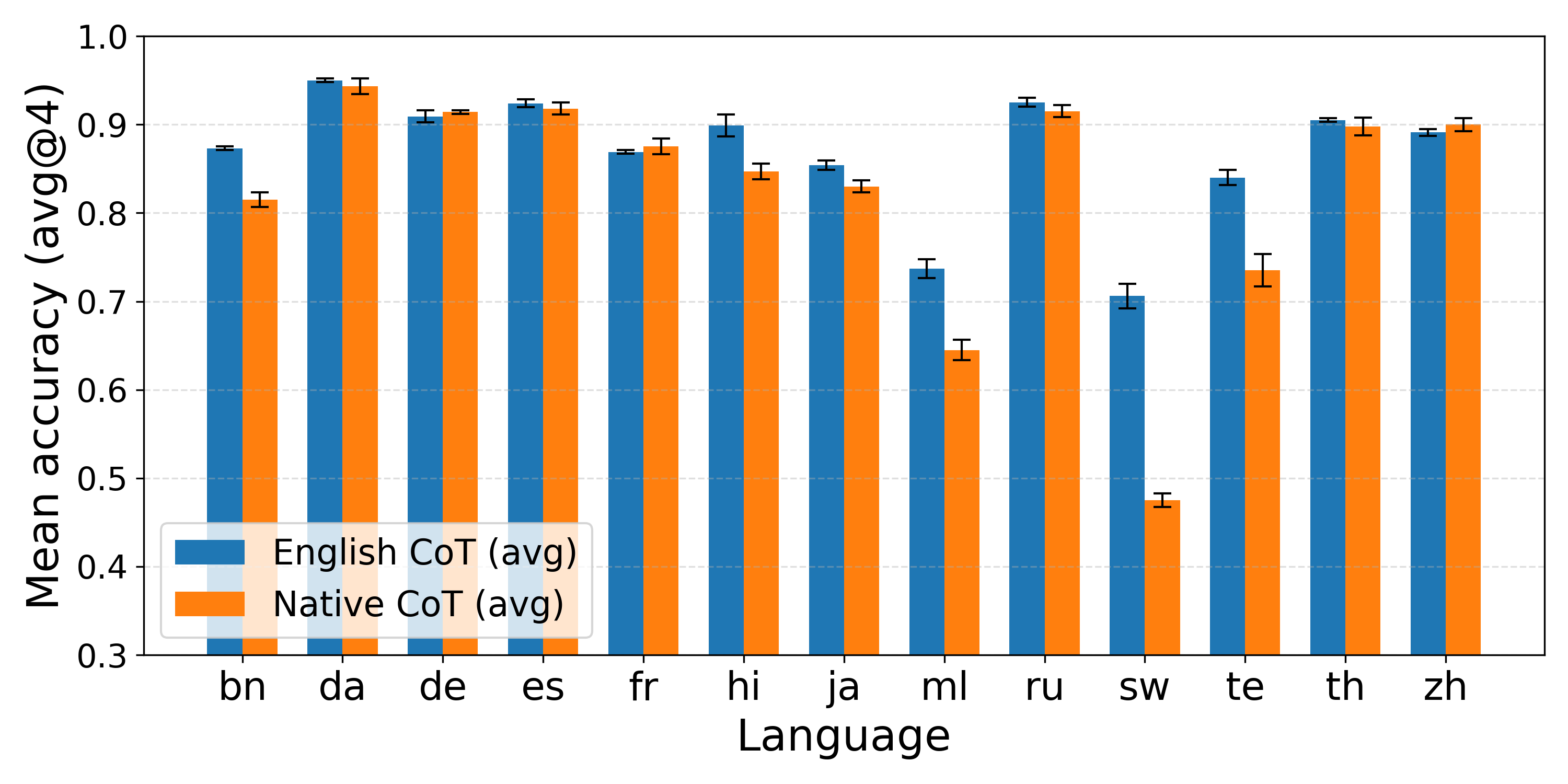} 
        \caption{MGSM}
        \label{fig:Qwen3-14B MGSM}
    
    \end{subfigure}
     \hfill
    \begin{subfigure}{0.48\textwidth} 
        \centering
        \includegraphics[width= \textwidth]{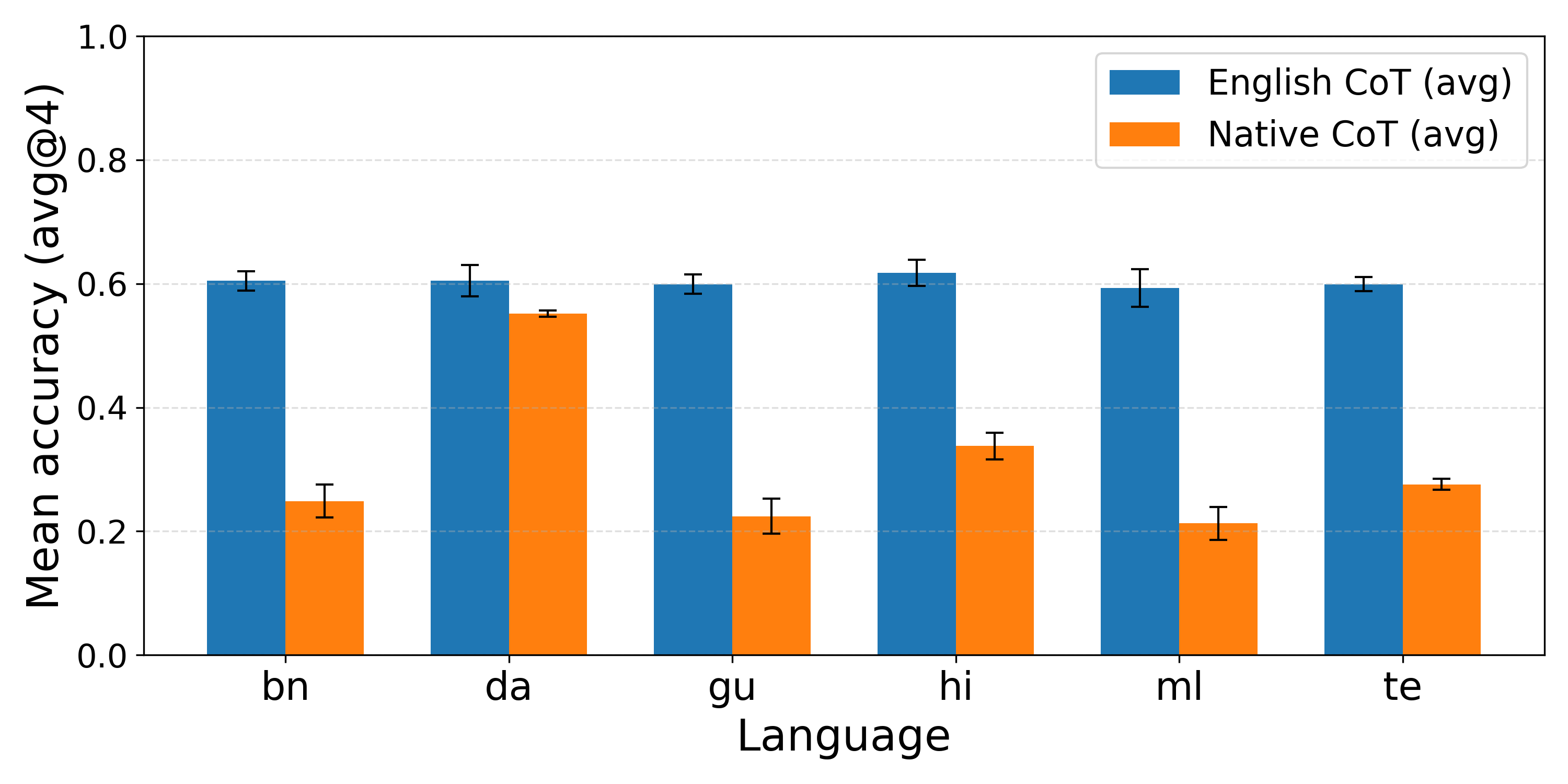} 
        \caption{GPQA Diamond}
        \label{fig:Qwen3-14B GPQA}
    \end{subfigure}
     \hfill
     \caption{\textbf{Reasoning in English vs Question's Language} is compared via final answer accuracy for \textbf{Qwen3 14B} for MGSM and GPQA diamond task. Error bars denote standard deviation.}
                
\label{fig:en_cot_vs_native_cot qwen3 14b}
\end{figure*}

\begin{figure*}[h!]
    \centering
        \hfill
    \begin{subfigure}{0.48\textwidth} 
        \centering
        \includegraphics[width= \textwidth]{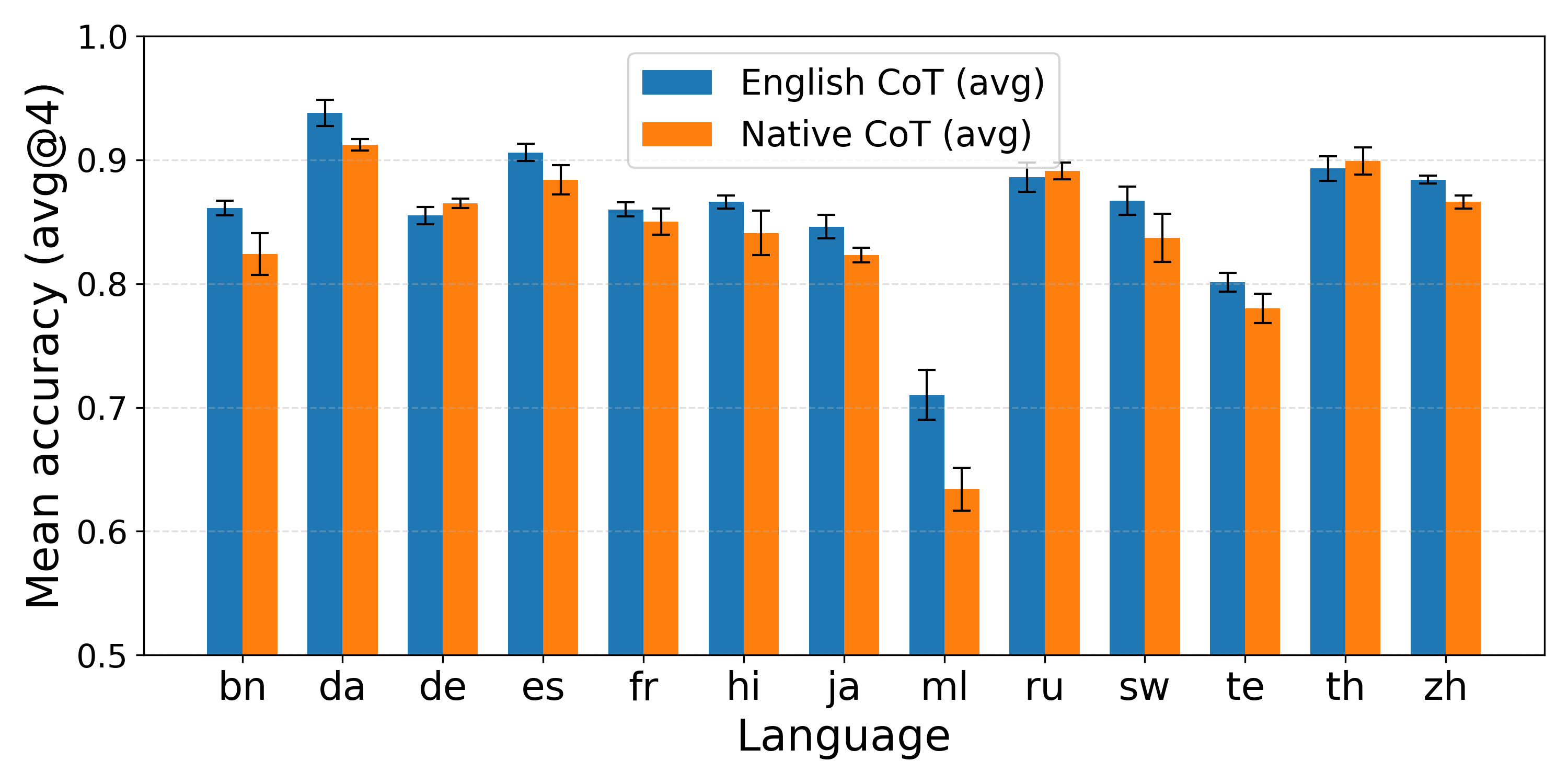} 
        \caption{MGSM}
        \label{fig:deepseek-ai_DeepSeek-R1-Distill-Llama-70B MGSM}
    
    \end{subfigure}
     \hfill
    \begin{subfigure}{0.48\textwidth} 
        \centering
        \includegraphics[width= \textwidth]{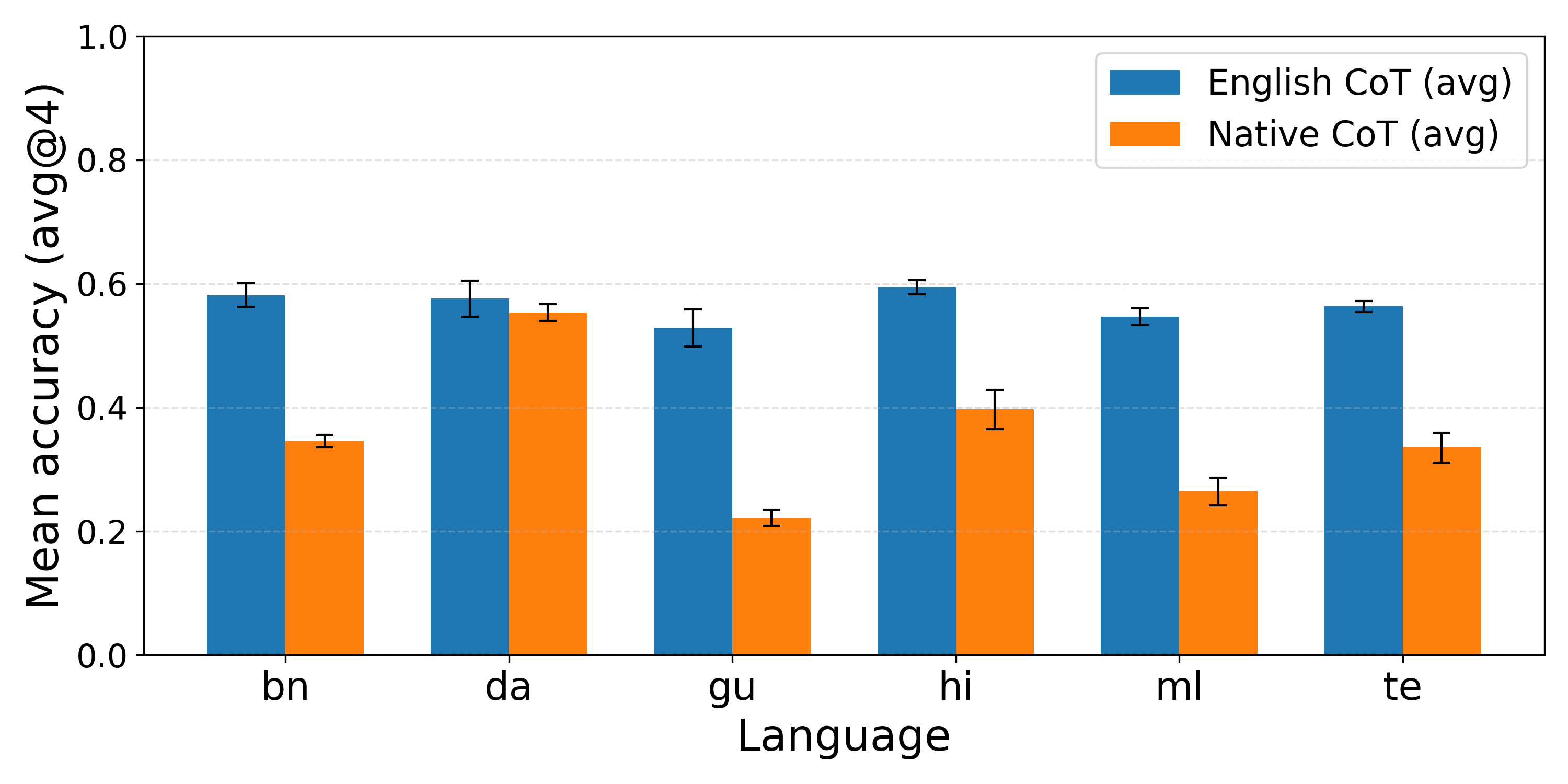} 
        \caption{GPQA Diamond}
        \label{fig:deepseek-ai_DeepSeek-R1-Distill-Llama-70B GPQA}
    \end{subfigure}
     \hfill
     \caption{\textbf{Reasoning in English vs Question's Language} is compared via final answer accuracy for \textbf{DeepSeek-R1-Distill-Llama-70B} for MGSM and GPQA diamond task. Error bars denote standard deviation.}
                
\label{fig:en_cot_vs_native_cot deepseek-ai_DeepSeek-R1-Distill-Llama-70B}
\end{figure*}

\begin{figure*}[h!]
    \centering
        \hfill
    \begin{subfigure}{0.48\textwidth} 
        \centering
        \includegraphics[width= \textwidth]{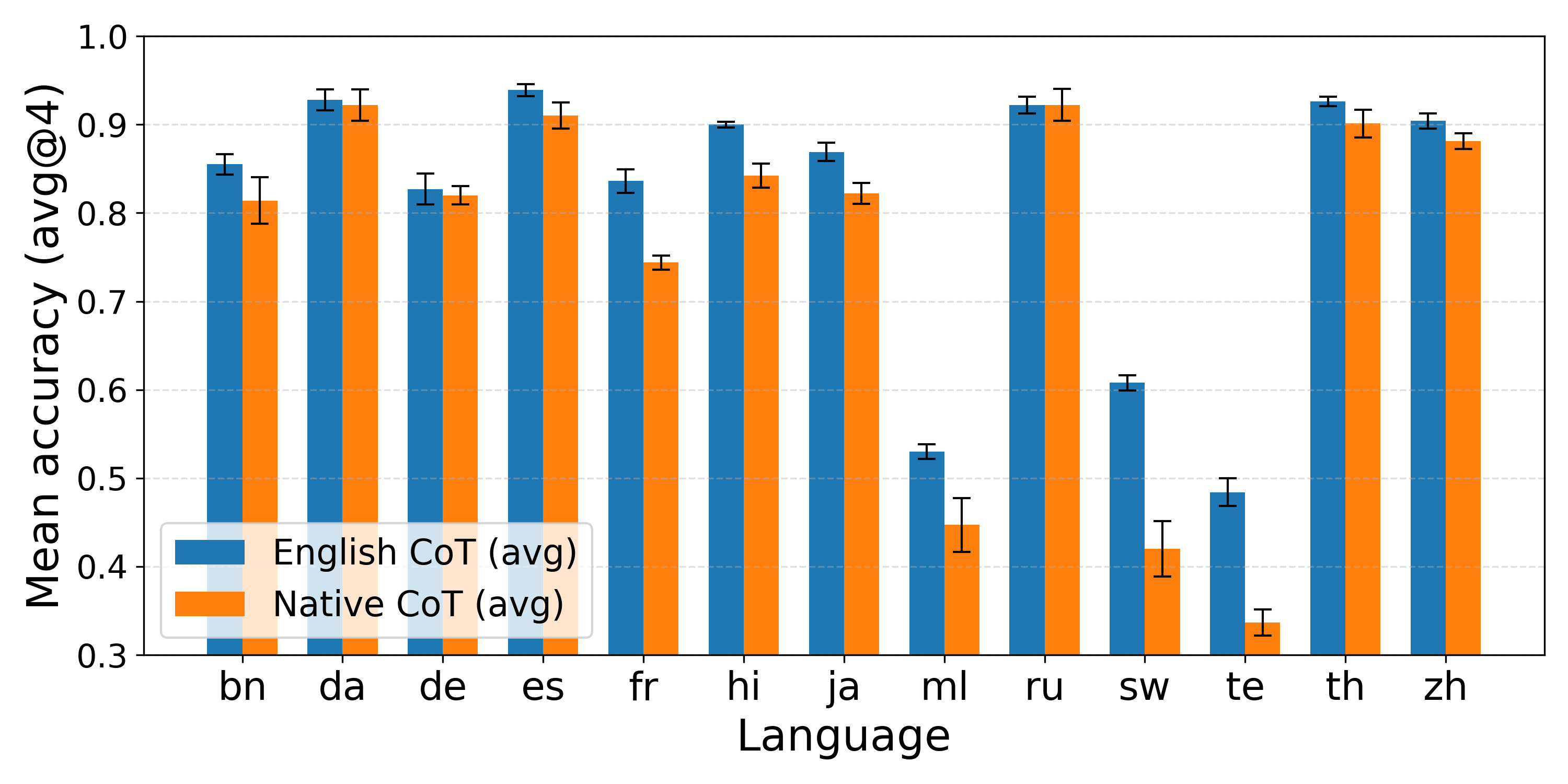} 
        \caption{MGSM}
        \label{fig:deepseek-ai_DeepSeek-R1-Distill-Qwen-32B MGSM}
    
    \end{subfigure}
     \hfill
    \begin{subfigure}{0.48\textwidth} 
        \centering
        \includegraphics[width= \textwidth]{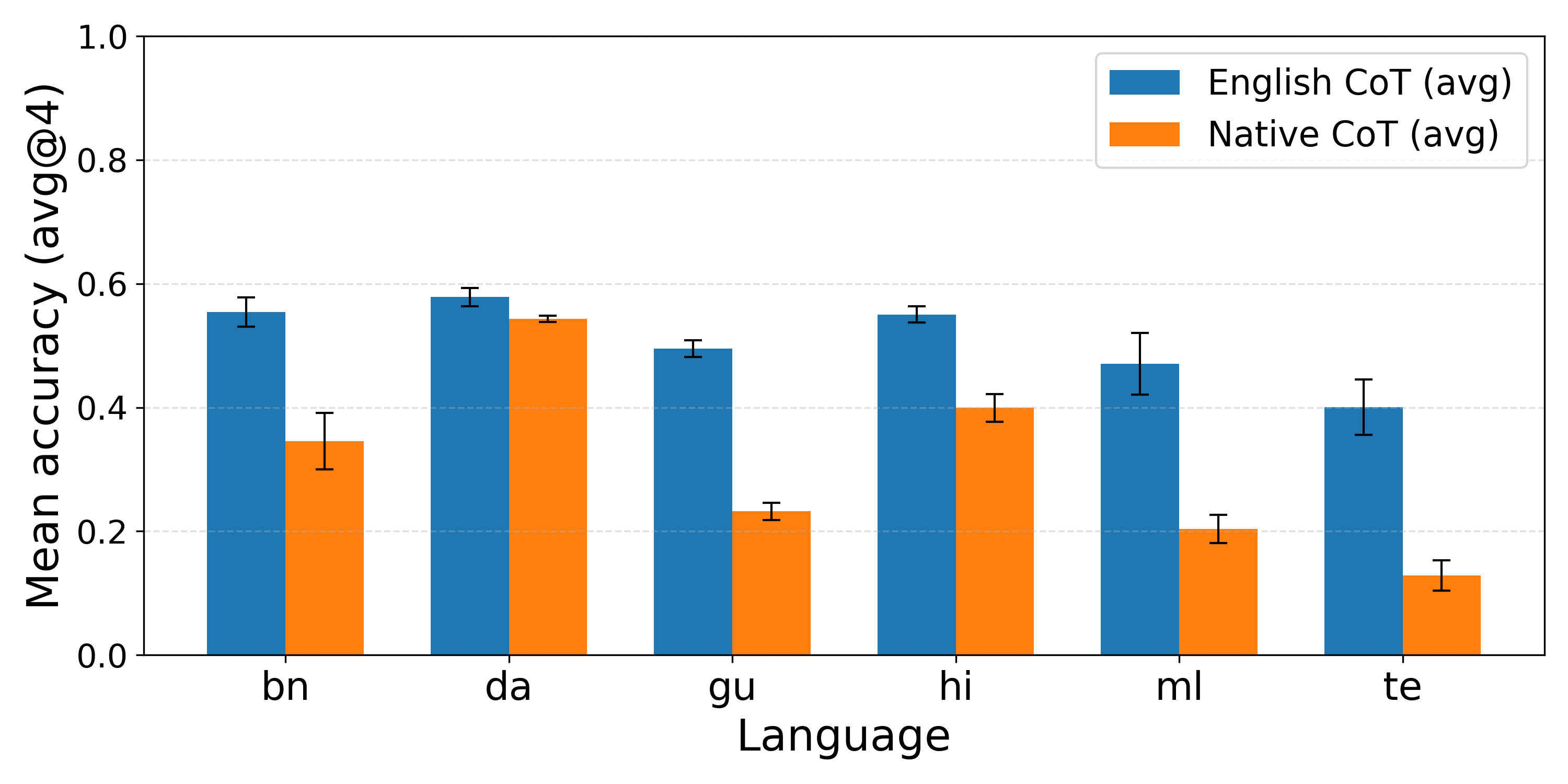} 
        \caption{GPQA Diamond}
        \label{fig:deepseek-ai_DeepSeek-R1-Distill-Qwen-32B GPQA}
    \end{subfigure}
     \hfill
     \caption{\textbf{Reasoning in English vs Question's Language} is compared via final answer accuracy for \textbf{DeepSeek-R1-Distill-Qwen-32B} for MGSM and GPQA diamond task. Error bars denote standard deviation.}
                
\label{fig:en_cot_vs_native_cot deepseek-ai_DeepSeek-R1-Distill-Qwen-32B}
\end{figure*}

Comparison of reasoning in English vs question's language for additional models can be found in Figures \ref{fig:en_cot_vs_native_cot qwen QwQ 32b} - \ref{fig:en_cot_vs_native_cot deepseek-ai_DeepSeek-R1-Distill-Qwen-32B}. These observations align with our findings for the Qwen QwQ-32B model; specifically, the performance gap between reasoning in English and in the question’s language for multilingual questions is substantially larger for GPQA Diamond than for MGSM.

\subsection{Standard deviation of accuracy for MGSM}
\label{sec: native_cot_reasoning_std}
In Figure \ref{fig:native_cot_reasoning_std} we present the standard deviation of accuracy of reasoning models on MGSM over 4 trials. The models reason in questions language for the multilingual task. Standard deviation remain below \textbf{0.05} in all instances.

\section{Analyzing Cognitive Behaviors: Additional details}
\label{sec: Analyzing Cognitive Behaviors: Additional details}

Table \ref{tab:cognitive_attributes} presents the prompt templates used to count the occurrence of cognitive behaviors in the reasoning trace when an LRM reasons in questions language. We replace "questions language" with "English" in the prompt when we analyze reasoning trace in English. Using gpt-4o-mini, we count the occurrences of these attributes by enclosing them within <count> and </count> tags, then compute the average across samples and subsequently across four trials.

\subsection{Human - LLM agreement study}
\label{sec: Analyzing Cognitive Behaviors: human lm agreement}
To validate the LLM-as-a-judge evaluator, we conducted a human annotation study for the cognitive behavior analysis task. We randomly sampled 30 evaluation prompts for each of the two low-resource languages - Hindi and  Malayalam, yielding 60 examples in total. Each example was independently solved and labeled by native-speaker annotators with strong grade-school mathematics skills. All annotators were computer science undergraduate students and each annotated 30 questions in their respective native language. Annotators were instructed to count and report the occurrence of each of the four cognitive behaviors in the reasoning trace using the same instructions given to the LLM evaluator. We compared the LLM judgments to the human labels and computed percent agreement and Cohen’s Kappa which are detailed in Table \ref{table:cognitive analysis human LM agreement}. The results indicate that the LLM evaluator delivers reasonably reliable assessments of cognitive behaviors, even for low-resource languages like Malayalam.

\section{Lost in Translation}
\label{sec: lost in translation}
In our experiments we investigated scenarios where reasoning in question's language outperformed reasoning in English. Across languages and across models, we observed that reasoning in question's language outperformed reasoning in English in scenarios where reasoning in English was incorrect due to errors introduced in translation. We refer to this as "Lost in Translation". 

\begin{figure*}[t]
  \centering
  \includegraphics[width=0.98\textwidth]{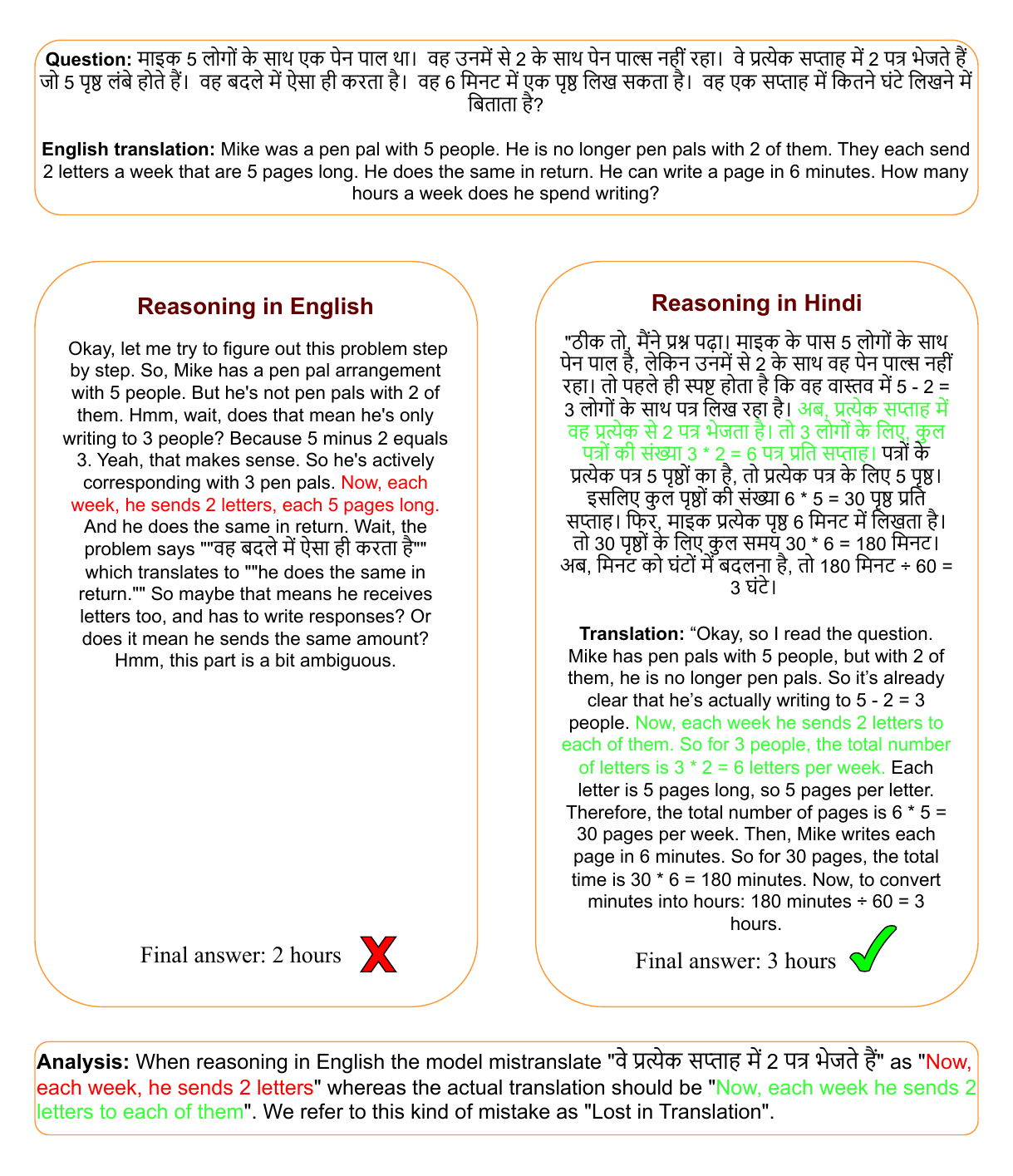}
    \caption{\textbf{Lost in Translation.} Here, we compare the reasoning traces in English and Hindi generated by Qwen QwQ-32B for a question originally written in Hindi. We observe that the English reasoning contains an error caused by a mistranslation (highlighted in red), while the corresponding correct sentence in the Hindi reasoning is highlighted in green. For the reader’s clarity, English translations of the Hindi question and reasoning are also provided.}
    \label{fig:LiT example}
\end{figure*}

Figure \ref{fig:LiT example} illustrates an example of Lost in Translation. Here, the English reasoning fails to arrive at the correct final answer due to a mistranslation. Specifically, the English reasoning interprets a key detail as “he sent two letters in total” instead of the correct meaning, “he sent two letters to each of them.” In contrast, the reasoning in the original question’s language does not encounter this issue, as no translation is involved, and therefore arrives at the correct answer.

\subsection{Lost in Translation - Quantitative Analysis}
\label{sec: lost in translation quantitative analysis}
We use gpt-4o-mini as an evaluator LLM, employing the prompt template shown in Table \ref{tab:lost_in_translation_simple} to determine whether an incorrect answer from reasoning in English occurred due to a mistranslation or not. From gpt-4o-mini’s output we extract a "Yes" or "No" enclosed in <classification></classification> tags, compute the fraction of incorrect answers attributable to mistranslation for each run, and report the mean value over four trials.

\subsection{Lost in Translation: Human-LLM agreement study}
\label{sec: lost in translation Human LM agreement}

To validate the LLM-as-a-judge evaluator, we conducted a human annotation study for "Lost in Translation" (LiT) task. We randomly sampled 50 LiT evaluation prompts from MGSM for each of three low-resource languages - Hindi, Bengali and  Malayalam, yielding 150 examples in total. Each example was independently solved and labeled by native-speaker annotators with strong grade-school mathematics skills. All annotators were computer science undergraduate students and each annotated 50 questions in their respective native language. Annotators were instructed to classify a reasoning trace as “Lost in Translation” or not based on the same instructions provided to the LLM judge. We compared the LLM judgments to the human labels and computed percent agreement, precision, recall, and Cohen’s Kappa which are detailed in Table \ref{table:LiT human LM agreement}. 
These results indicate that the LLM evaluator produces reasonably reliable evaluations for the LiT task even for low resource languages like Malayalam.

\begin{table*}[h]
    \centering
    \begin{tabular}{|l|c|c|c|c|}
        \hline
        \multirow{2}{*}{\textbf{Cognitive Behavior}} & \multicolumn{2}{c|}{\textbf{Hindi (30 samples)}} & \multicolumn{2}{c|}{\textbf{Malayalam (30 samples)}} \\ 
        \cline{2-5}
         & \textbf{\% Agreement} & \textbf{Kappa} & \textbf{\% Agreement} & \textbf{Kappa} \\ 
        \hline
        Subgoal & 0.73 & 0.68 & 0.67 & 0.62 \\ 
        \hline
        Verification & 0.80 & 0.69 & 0.70 & 0.58 \\ 
        \hline
        Backtracking & 0.80 & 0.58 & 0.733 & 0.54 \\ 
        \hline
        Backward Chaining & 0.83 & 0.52 & 0.87 & 0.52 \\ 
        \hline
    \end{tabular}
    \caption{ Human-LLM agreement for cognitive analysis task. Percent agreement and Cohen's Kappa are reported for Hindi and Malayalam.}
    \label{table:cognitive analysis human LM agreement}
\end{table*}

\begin{table*}[h!]
\centering

\begin{tabular}{l >{\raggedright\arraybackslash}p{0.75\textwidth}}
\toprule
\textbf{Cognitive Attribute} & \textbf{Prompt} \\
\midrule

\textbf{Verification} & 
Here is a chain-of-reasoning in the question's language that a Language Model generated for the question:

"\textit{\{question\}}"

The model’s reasoning (between \textless think\textgreater…\textless/think\textgreater) is:

\textit{\{chain\_of\_reasoning\}}

Evaluate whether this chain-of-reasoning contains any explicit answer-verification steps. An answer-verification step is any instance where the model checks its intermediate numeric result and asks itself if the answer is correct or not and probably goes on to re-check it. If you find any of these instances, count them and put the number between \textless count\textgreater and \textless/count\textgreater. Otherwise, output \textless count\textgreater0\textless/count\textgreater. \\

\midrule

\textbf{Backtracking} &
Here is a chain-of-reasoning in the question's language '\textit{\{L\}}' that a Language Model generated for the question: 

"\textit{\{question\}}"

The model’s reasoning (between \textless think\textgreater…\textless/think\textgreater) is:

\textit{\{chain\_of\_reasoning\}}

Evaluate whether this reasoning contains any backtracking behavior—i.e., places where the model decides that its previous approach won’t reach the correct answer and explicitly abandons that path, starting fresh on an alternative intermediate step. Count the number of such backtracking instances and put the result between \textless count\textgreater and \textless/count\textgreater. If none, output \textless count\textgreater0\textless/count\textgreater. \\

\midrule

\textbf{Subgoal setting} &
Here is a chain-of-reasoning in the question's language '\textit{\{L\}}' that a Language Model generated for the question: 

"\textit{\{question\}}"

The model’s reasoning (between \textless think\textgreater…\textless/think\textgreater) is:

\textit{\{chain\_of\_reasoning\}}

Evaluate whether this reasoning explicitly sets any subgoals (e.g., “First I will try to isolate x…”, “Next I aim to simplify …”, etc.) on the way toward \textit{\{target\}}. Count how many distinct subgoals appear and put that number between \textless count\textgreater and \textless/count\textgreater. If none, output \textless count\textgreater0\textless/count\textgreater. \\

\midrule

\textbf{Backward chaining} &
Here is a chain-of-reasoning in the question's language '\textit{\{L\}}' that a Language Model generated for the question: 

"\textit{\{question\}}"

The model’s reasoning (between \textless think\textgreater…\textless/think\textgreater) is:

\textit{\{chain\_of\_reasoning\}}

Evaluate whether this reasoning uses backward-chaining—i.e., it starts from the final answer and works backward to earlier steps. Count how many distinct backward-chaining instances occur and put that number between \textless count\textgreater and \textless/count\textgreater. If none, output \textless count\textgreater0\textless/count\textgreater. \\

\bottomrule
\end{tabular}
\caption{Cognitive Attributes Analysis Prompt Templates}
\label{tab:cognitive_attributes}
\end{table*}

\begin{table*}[ht]
  \centering  
  \begin{tabular}{lcccc}
    \toprule
     \textbf{Language (Samples)}
    & \textbf{Percent agreement} & \textbf{Precision} & \textbf{Recall} & \textbf{Cohen's kappa} \\
    \midrule
{}    Hindi (50)     & 0.78 & 0.83 & 0.80 & 0.55 \\
    Malayalam (50)  & 0.76 & 0.79 & 0.79 & 0.50 \\
    Bengali (50)    & 0.78 & 0.72 & 0.82 & 0.56 \\
    \midrule
    \textbf{Total (150)} & \textbf{0.77} & \textbf{0.78} & \textbf{0.80} & \textbf{0.54} \\
    \bottomrule
  \end{tabular}
  \caption{Lost in Translation: Human–LLM agreement across 150 samples (50 per language).}
  \label{table:LiT human LM agreement}
\end{table*}


\begin{table*}[t!]
\centering

\begin{tabularx}{\textwidth}{lX}
\toprule
\textbf{Task} & \textbf{Prompt Template} \\
\midrule
MGSM/ GPQA Diamond & 
Here is a chain-of-reasoning that a Language Model generated for the question: \par
\texttt{"\{question\}"} \par \medskip
The model’s reasoning (between \texttt{<think>...</think>}) is: \par
\texttt{\{chain\_of\_reasoning\}} \par \medskip
The chain-of-reasoning here has led to incorrect answer. Here the model reasons in English, check if here some information available in the question has been omitted/ mis-translated while translating the question to English. If so output 'Yes' else 'No". Put the Output in between \texttt{<classification>} and \texttt{</classification>} tags. Do not generate anything else. \\
\bottomrule
\end{tabularx}

\caption{Lost in Translation - Prompt templates}
\label{tab:lost_in_translation_simple}
\end{table*}



\end{document}